\documentclass{article} 
\usepackage{icais2025_conference,times}

\usepackage{amsmath,amsfonts,bm}

\def\eqref#1{equation~\ref{#1}}

\def\1{\bm{1}}

\DeclareMathAlphabet{\mathsfit}{\encodingdefault}{\sfdefault}{m}{sl}
\SetMathAlphabet{\mathsfit}{bold}{\encodingdefault}{\sfdefault}{bx}{n}

\usepackage[utf8]{inputenc}
\usepackage{amsmath,amssymb,amsthm}
\usepackage{graphicx}
\usepackage{hyperref}
\usepackage{natbib}
\usepackage{geometry}
\usepackage{booktabs}
\usepackage{multirow}
\usepackage{algorithm}
\usepackage{algorithmic}
\usepackage{xcolor}

\usepackage{tikz}
\usepackage{amsmath} 
\usetikzlibrary{
    shapes.geometric, 
    arrows.meta,      
    positioning,      
    calc              
}

\newtheorem{theorem}{Theorem}

\newtheorem{definition}{Definition}

\newcommand{\RR}{\mathbb{R}}

\newcommand{\CC}{\mathbb{C}}
\newcommand{\norm}[1]{\left\|#1\right\|}
\newcommand{\inner}[2]{\left\langle#1,#2\right\rangle}

\definecolor{primaryblue}{RGB}{52, 152, 219}
\definecolor{secondarygreen}{RGB}{46, 204, 113}
\definecolor{accentorange}{RGB}{230, 126, 34}
\definecolor{deepred}{RGB}{192, 57, 43}
\definecolor{softpurple}{RGB}{155, 89, 182}
\definecolor{lightgray}{RGB}{236, 240, 241}
\definecolor{darkgray}{RGB}{52, 73, 94}
\usepackage{hyperref}
\hypersetup{
    colorlinks=true,
    linkcolor=blue!70!black,
    citecolor=green!60!black,
    urlcolor=red!60!black
}
\usepackage{url}

\title{\LARGE\bf Physics-Informed Neural Networks and Neural Operators for Parametric PDEs\thanks{While artificial intelligence (AI) demonstrates significant potential for accelerating scientific research, this work represents, to our knowledge, the first AI-collaborative survey in the AI for PDEs domain. It has been submitted to the AI Track (Research Generated by AI Systems) of The 1st International Conference on AI Scientists (ICAIS 2025). This paper was developed through a specific human-AI iterative process: the human authors provided instructions and research direction. The core content generation was performed by Claude Sonnet 4.5, with Gemini 2.5 Pro and GPT 5 used for refining prompt templates and formatting. A key challenge addressed was large model hallucination. To mitigate this, our workflow involved compelling the AI to meticulously cross-check references and return accessible links to the full-text literature, which enabled a final, thorough human verification. Given the rapid evolution of the field and the limitations of this process, we present this as an evolving AI-generated survey and intend to incrementally update it in future versions.}}

\author{Zhuo Zhang\\
College of Computer Science and Technology \\
National University of Defense Technology \\
Changsha, Hunan, China \\
\texttt{zhangzhuo@nudt.edu.cn} \\
\And
Xiong Xiong \\
School of Mathematics and Statistics \\
Northwestern Polytechnical University \\
Xi'an, Shaanxi, China \\
\texttt{xiongxiongwpu@mail.nwpu.edu.cn} \\
\And
Sen Zhang \\
College of Computer Science and Technology \\
National University of Defense Technology \\
Changsha, Hunan, China \\
\texttt{zhangsen19@nudt.edu.cn} \\
\And
Yuan Zhao\\
College of Computer Science and Technology \\
National University of Defense Technology \\
Changsha, Hunan, China \\
\texttt{zhaoyuan@nudt.edu.cn} \\
\And
Xi Yang \\ 
  College of Computer Science and Technology \\
  National Key Laboratory of Parallel and Distributed Computing \\
  National University of Defense Technology \\
  Changsha, Hunan, China \\
  \texttt{yangxi1016@nudt.edu.cn}
}

\iclrfinalcopy 
\begin{document}

\maketitle

\begin{figure}[H]
\centering
\includegraphics[width=\textwidth]{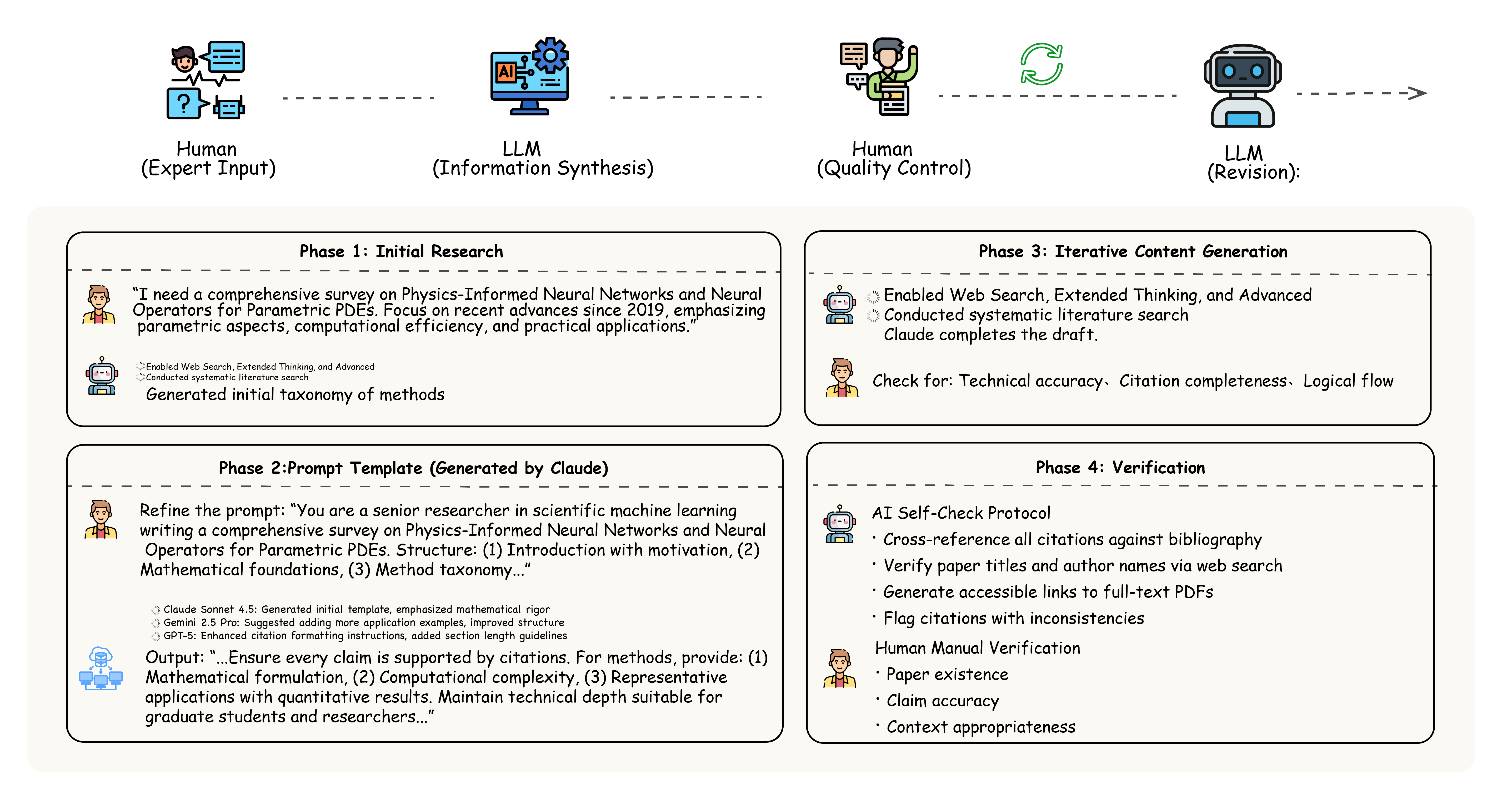}  \vspace{-0.5cm}

\caption{Human-AI collaborative workflow for survey generation.}
\label{fig:collaboration_workflow}
\end{figure}

\begin{abstract}
As artificial intelligence evolves from an assistive tool into an agent capable of independent or collaborative scientific discovery, new research paradigms are emerging. This work serves as an example of this paradigm, presenting a comprehensive technical overview of parametric Parametric partial differential equations (PDEs) solvers generated through an iterative human-AI interactive process.

PDEs arise ubiquitously in science and engineering, where solutions depend on parameters representing physical properties, boundary conditions, or geometric configurations. Traditional numerical methods require solving the PDE anew for each parameter value, making parameter space exploration prohibitively expensive for high-dimensional problems. Recent advances in machine learning, particularly physics-informed neural networks (PINNs) and neural operators, have revolutionized parametric PDE solving by learning solution operators that generalize across parameter spaces. We critically analyze two main paradigms: (1) PINNs, which embed physical laws as soft constraints and excel at inverse problems with sparse data, and (2) neural operators (including DeepONet, Fourier Neural Operator, and their variants), which learn mappings between infinite-dimensional function spaces and achieve unprecedented parameter space generalization. Through detailed comparisons across fluid dynamics, solid mechanics, heat transfer, and electromagnetics, we show that neural operators can achieve computational speedups ranging from $10^3$ to $10^5$ times faster than traditional solvers for multi-query scenarios, while maintaining comparable accuracy. We provide practical guidance for method selection, discuss theoretical foundations including universal approximation and convergence guarantees, and identify critical open challenges including high-dimensional parameter spaces, complex geometries, and out-of-distribution generalization. This work establishes a unified framework for understanding parametric PDE solvers through the lens of operator learning, offering a comprehensive resource—which we intend to incrementally update—for this rapidly evolving field. 

\end{abstract}
\section*{Nomenclature}
\addcontentsline{toc}{section}{Nomenclature}

\begin{table}[H]
\centering
\begin{tabular}{ll}
\toprule
\textbf{Symbol} & \textbf{Meaning} \\
\midrule
$\mu$ & Parameter vector \\
$\mathcal{P}$ & Parameter space \\
$d$ & Parameter dimension \\
$\Omega$ & Spatial domain \\
$u(x,t;\mu)$ & Parameter-dependent PDE solution \\
$\mathcal{L}(\cdot;\mu)$ & Parameter-dependent differential operator \\
$\mathcal{G}$ & Solution operator mapping parameters to solutions \\
$\mathcal{A}$ & Input function space \\
$\mathcal{U}$ & Solution function space \\
$\theta$ & Neural network parameters \\
$\mathcal{L}_{\text{PDE}}$ & Physics-informed loss \\
$\mathcal{L}_{\text{data}}$ & Data fidelity loss \\
Re & Reynolds number \\
DOF & Degrees of freedom \\
\bottomrule
\end{tabular}
\end{table}


\textbf{Note on Methodology:} This survey represents an exploratory effort in AI-assisted scientific writing. As illustrated in Figure~\ref{fig:collaboration_workflow}, the content was generated through a structured human-AI collaboration involving systematic literature search and multi-round revision.

\section{Introduction and Background}
\label{sec:introduction}

\subsection{Motivation: The Parametric Challenge in Scientific Computing}
\label{sec:motivation}

Parametric partial differential equations (PDEs) represent one of the most fundamental yet computationally challenging problems in scientific computing and engineering. These equations, whose solutions depend on parameters representing physical properties, boundary conditions, or geometric configurations, arise ubiquitously across disciplines: from material design requiring exploration of compositional parameter spaces, to fluid dynamics demanding Reynolds number sweeps, to uncertainty quantification necessitating sampling over parameter distributions.

Consider a canonical example from computational fluid dynamics: the flow around an airfoil depends critically on parameters such as the Reynolds number, angle of attack, and airfoil shape. Traditional computational fluid dynamics (CFD) solvers, while mature and reliable, must solve the governing Navier-Stokes equations independently for each parameter configuration. For a modest parameter space exploration involving just three parameters with 10 values each, this requires 1,000 independent simulations—a prohibitively expensive endeavor when each simulation demands hours or days of computation on high-performance computing clusters.

This ``one-parameter-at-a-time" limitation of traditional numerical methods becomes even more severe in several critical scenarios:

\textbf{Real-time decision making:} In applications such as autonomous systems, medical diagnostics, or process control, solutions must be obtained in milliseconds rather than hours. For instance, patient-specific cardiovascular simulations for surgical planning cannot wait for overnight cluster computations.

\textbf{Inverse problems and parameter identification:} Determining unknown physical parameters from observational data—such as inferring material properties from experimental measurements or identifying hidden forcing terms in climate models—requires repeatedly solving forward problems with different parameter guesses. Traditional optimization loops involving thousands of PDE solves become computationally intractable.

\textbf{Uncertainty quantification (UQ):} Modern engineering design must account for uncertainties in material properties, manufacturing tolerances, and operational conditions. Monte Carlo methods for UQ typically require $10^4$-$10^6$ PDE evaluations to accurately estimate probability distributions—a task infeasible with conventional solvers for complex systems.

\textbf{High-dimensional parameter spaces:} Many real-world problems involve tens or hundreds of parameters. For example, weather forecasting models contain hundreds of thousands of uncertain parameters, molecular dynamics simulations depend on high-dimensional potential energy surfaces, and topology optimization problems search over spaces of possible geometric configurations. Traditional methods suffer from the curse of dimensionality, where sampling complexity grows exponentially with parameter dimension.

The scientific and economic implications are profound. The aviation industry spends billions on wind tunnel testing and CFD simulations for aircraft design optimization. The pharmaceutical sector requires years of computational protein folding studies for drug discovery. Climate modeling centers consume enormous computational resources for ensemble forecasts that still struggle to quantify uncertainties adequately.

\textbf{The Central Thesis:} While parametric PDEs are ubiquitous in science and engineering, traditional numerical methods solve one parameter configuration at a time, making parameter space exploration prohibitively expensive. This survey examines how machine learning methods—particularly physics-informed neural networks and neural operators—are transforming our ability to solve parametric PDEs by learning solution operators that generalize across entire parameter spaces. For multi-query scenarios where solutions are needed at many parameter values, these methods enable computational speedups of $10^3$ to $10^5$ times compared to traditional solvers, with documented examples including 60,000$\times$ speedup for turbulence \citep{Li2021ICLR} and 1,000$\times$ speedup for laminar flows \citep{Jin2021}.

\subsection{Mathematical Formulation of Parametric PDEs}
\label{sec:math_formulation}

To establish precise terminology and mathematical foundations, we formally define the class of parametric PDEs that forms the focus of this survey.

\begin{definition}[Parametric PDE]
Let $\mathcal{P} \subset \RR^d$ denote a parameter space of dimension $d$. A parametric PDE is defined as:
\begin{equation}
\mathcal{L}(u(x,t;\mu); \mu) = f(x,t;\mu), \quad \forall \mu \in \mathcal{P}, \quad (x,t) \in \Omega \times [0,T],
\label{eq:parametric_pde}
\end{equation}
equipped with parameter-dependent boundary and initial conditions:
\begin{align}
\mathcal{B}(u(x,t;\mu); \mu) &= g(x,t;\mu), \quad (x,t) \in \partial\Omega \times [0,T], \\
u(x,0;\mu) &= u_0(x;\mu), \quad x \in \Omega,
\end{align}
where $u: \Omega \times [0,T] \times \mathcal{P} \to \RR^n$ is the parameter-dependent solution (possibly vector-valued), $\mathcal{L}(\cdot;\mu)$ is a parameter-dependent differential operator, $\mathcal{B}(\cdot;\mu)$ specifies boundary conditions, and $f, g, u_0$ represent parameter-dependent forcing, boundary, and initial data.
\end{definition}

This formulation encompasses a vast range of problems. The parameter $\mu$ can represent:

\textbf{1. Physical parameters:} Material properties (viscosity, thermal conductivity, elastic moduli), flow regime indicators (Reynolds, Rayleigh, Péclet numbers), or forcing magnitudes.

\textit{Example:} The parameterized Burgers equation models nonlinear wave propagation with viscosity parameter $\nu$:
\begin{equation}
\frac{\partial u}{\partial t} + u\frac{\partial u}{\partial x} = \nu \frac{\partial^2 u}{\partial x^2}, \quad \nu \in [\nu_{\min}, \nu_{\max}] \subset \RR_+
\end{equation}
Here $\mathcal{P} = [\nu_{\min}, \nu_{\max}]$ represents the one-dimensional parameter space of viscosities.

\textbf{2. Geometric parameters:} Domain shapes, boundary configurations, or obstacle positions.

\textit{Example:} Flow around parameterized airfoils, where $\mu$ represents coefficients in a geometric parameterization (e.g., NACA profile parameters or Bézier control points):
\begin{equation}
\begin{cases}
-\nabla \cdot (\nu \nabla u) + (u \cdot \nabla)u + \nabla p = 0 & \text{in } \Omega(\mu) \\
\nabla \cdot u = 0 & \text{in } \Omega(\mu) \\
u = u_{\infty} & \text{on } \partial\Omega_{\text{inlet}} \\
u = 0 & \text{on } \partial\Omega_{\text{airfoil}}(\mu)
\end{cases}
\end{equation}
where the domain $\Omega(\mu)$ and airfoil boundary $\partial\Omega_{\text{airfoil}}(\mu)$ depend explicitly on shape parameters.

\textbf{3. Boundary and initial condition parameters:} Inlet velocities, temperature distributions, loading patterns, or initial state configurations.

\textit{Example:} Heat conduction with parameterized boundary temperature:
\begin{equation}
\begin{cases}
\rho c_p \frac{\partial T}{\partial t} = \nabla \cdot (k \nabla T) & \text{in } \Omega \\
T = T_{\text{wall}}(\mu) & \text{on } \partial\Omega
\end{cases}
\end{equation}
where $\mu$ parameterizes the boundary temperature profile.

\textbf{4. Source term parameters:} External forcing magnitudes, distributions, or time-dependent profiles.

The \textbf{solution manifold} is a central concept:
\begin{equation}
\mathcal{M} = \{u(\cdot,\cdot;\mu) : \mu \in \mathcal{P}\} \subset \mathcal{U}
\end{equation}
where $\mathcal{U}$ is an appropriate function space (e.g., Sobolev space). The manifold $\mathcal{M}$ represents the set of all possible solutions as parameters vary. Understanding the geometric and topological properties of $\mathcal{M}$ is crucial for developing efficient approximation methods.

From an operator perspective, we seek to approximate the \textbf{parameter-to-solution map} (or solution operator):
\begin{equation}
\mathcal{G}: \mathcal{P} \to \mathcal{U}, \quad \mu \mapsto u(\cdot,\cdot;\mu)
\label{eq:solution_operator}
\end{equation}
This operator encodes the complete input-output relationship of the parametric PDE. Traditional numerical methods construct $\mathcal{G}$ implicitly by solving equation \eqref{eq:parametric_pde} for each query $\mu$. In contrast, machine learning approaches aim to learn an explicit approximation $\mathcal{G}_\theta \approx \mathcal{G}$ that can be rapidly evaluated.

\textbf{Key Challenges in Parametric PDEs:}

\textit{High-dimensional parameter spaces:} When $d \gg 1$, the curse of dimensionality emerges. Uniform sampling of a $d$-dimensional space with $n$ points per dimension requires $n^d$ samples—exponentially expensive. Even advanced techniques like sparse grids or reduced basis methods face limitations beyond $d \sim 20$.

\textit{Nonlinear parameter dependence:} The solution $u(\cdot,\cdot;\mu)$ may depend nonlinearly, non-smoothly, or even discontinuously on $\mu$. For instance, solutions may exhibit bifurcations, phase transitions, or shock formations at critical parameter values, making smooth interpolation impossible.

\textit{Multi-scale phenomena:} Parameters often control multiple length or time scales simultaneously. A small change in Reynolds number can trigger transition from laminar to turbulent flow, introducing fine-scale structures that traditional coarse approximations cannot capture.

\textit{Geometry-parameter coupling:} When geometry itself is parameterized, standard grid-based methods struggle. Each parameter value may require mesh regeneration, and solutions on different meshes cannot be directly compared or interpolated.

The mathematical challenge can be formally stated: \textit{Given finite computational resources and training data $\{(\mu_i, u_i)\}_{i=1}^N$ where $u_i = u(\cdot,\cdot;\mu_i)$, construct an approximation $\mathcal{G}_\theta$ such that $\norm{\mathcal{G}(\mu) - \mathcal{G}_\theta(\mu)}_{\mathcal{U}} < \epsilon$ for all $\mu \in \mathcal{P}$, with rapid evaluation $\mathcal{O}(1)$ ms per query.}

\subsection{Traditional Approaches: Reduced Order Models}
\label{sec:traditional_rom}

Before examining modern machine learning methods, we review traditional approaches to parametric PDEs, which provide both historical context and performance baselines. Classical reduced order models (ROMs) have been developed over decades, establishing rigorous mathematical foundations that inform contemporary developments.

\subsubsection{Proper Orthogonal Decomposition and Galerkin Projection}

The Proper Orthogonal Decomposition (POD), also known as Principal Component Analysis or Karhunen-Loève expansion, forms the cornerstone of traditional ROM approaches \citep{Holmes1996, Berkooz1993}. Given snapshot solutions $\{u(\cdot,\cdot;\mu_i)\}_{i=1}^N$ collected at training parameters, POD constructs an optimal low-dimensional approximation subspace.

The method computes spatial modes $\{\phi_k\}_{k=1}^{N_{\text{POD}}}$ by solving the eigenvalue problem:
\begin{equation}
C\phi_k = \lambda_k \phi_k, \quad C_{ij} = \inner{u_i}{u_j}_{\mathcal{U}}
\end{equation}
where $\inner{\cdot}{\cdot}_{\mathcal{U}}$ denotes an appropriate inner product. Solutions for new parameters are approximated as:
\begin{equation}
u(x,t;\mu) \approx \sum_{k=1}^{N_{\text{POD}}} a_k(t;\mu) \phi_k(x)
\end{equation}
where coefficients $a_k(t;\mu)$ are determined by Galerkin projection of the governing PDE onto the reduced space.

\textbf{Strengths:} POD-Galerkin excels when solutions lie near a low-dimensional linear subspace, achieving exponential convergence for smooth parameter dependence. Online evaluation is extremely fast ($\mathcal{O}(\text{ms})$) once modes are computed.

\textbf{Limitations for Parametric Problems:}
\textit{Linear assumption:} POD constructs a linear subspace, failing when the solution manifold $\mathcal{M}$ is intrinsically nonlinear (as in transport-dominated or bifurcating systems)., \textit{Parameter-dependent training:} Modes depend on the specific parameter samples chosen. Poor parameter space coverage leads to inadequate approximations., and \textit{Dimensionality limits:} While effective for $d < 10$, performance degrades rapidly in higher dimensions without adaptive sampling.

\subsubsection{Reduced Basis Methods}

Reduced basis (RB) methods \citep{Quarteroni2015, Hesthaven2016, Rozza2008} extend POD through sophisticated greedy algorithms for parameter space exploration. The key innovation is adaptive selection of snapshot parameters to maximize approximation quality across $\mathcal{P}$.

The greedy algorithm iteratively identifies the parameter $\mu^*$ where the current reduced model has largest error:
\begin{equation}
\mu^{n+1} = \arg\max_{\mu \in \mathcal{P}} \eta(\mu; \{\phi_k\}_{k=1}^n)
\end{equation}
where $\eta(\mu;\cdot)$ is a computable error estimator. A new solution snapshot at $\mu^{n+1}$ is computed and orthogonalized against existing basis functions.

\textbf{Strengths:} Rigorous a posteriori error bounds, systematic parameter space coverage, and theoretical optimality guarantees under affine parameter dependence.

\textbf{Limitations:} Despite these advantages, RB methods face several constraints. First, they require affine parameter dependence, meaning PDE operators must decompose as $\mathcal{L}(\cdot;\mu) = \sum_{q=1}^Q \Theta_q(\mu) \mathcal{L}_q(\cdot)$ with parameter-independent operators $\mathcal{L}_q$. This requirement excludes geometry-parameterized problems and many nonlinear cases. Second, the greedy procedure incurs substantial offline computational expense, requiring dozens to hundreds of full-order model solves. Third, like POD, RB methods face challenges beyond $d \sim 20$ despite sophisticated sampling strategies.

\subsubsection{Sparse Grid and Multi-Level Methods}

For higher-dimensional parameter spaces, sparse grid methods based on Smolyak's algorithm \citep{Bungartz2004} provide a middle ground between full tensor-product grids and random sampling. The key idea is anisotropic refinement, placing more collocation points along directions of high solution variability.

Multi-level Monte Carlo (MLMC) \citep{Giles2015} tackles high-dimensional uncertainty quantification by combining low-fidelity (coarse mesh) solutions with high-fidelity corrections, achieving variance reduction without prohibitive cost.

\textbf{Limitations:} Sparse grids require smoothness assumptions and struggle beyond $d \sim 30$. MLMC helps with UQ but does not provide a general-purpose surrogate for $\mathcal{G}$.

\subsubsection{Comparative Performance Analysis}

Table \ref{tab:traditional_methods} summarizes the capabilities and limitations of traditional ROM approaches.

\begin{table}[h]
\centering
\caption{Comparison of traditional reduced order modeling approaches for parametric PDEs}
\label{tab:traditional_methods}
\small
\begin{tabular}{p{3cm}p{2cm}p{2cm}p{2.5cm}p{4cm}}
\toprule
\textbf{Method} & \textbf{Parameter Dimension} & \textbf{Online Query Speed} & \textbf{Offline Training Cost} & \textbf{Main Limitations} \\
\midrule
POD-Galerkin & Low ($d<10$) & Fast ($\sim$ms) & Moderate (snapshot collection) & Linear subspace assumption; parameter coverage dependency \\
\midrule
Reduced Basis & Low ($d<10$) & Very Fast ($\sim$ms) & High (greedy iterations) & Affine parameter dependence; geometry limitations \\
\midrule
Sparse Grids & Medium ($d<20-30$) & Moderate & High (collocation points) & Smoothness requirements; curse of dimensionality persists \\
\midrule
Multi-Level MC & High ($d>50$) & N/A (statistical) & Very High & UQ-specific; no surrogate model \\
\bottomrule
\end{tabular}
\end{table}

\textbf{The Gap and Opportunity:} Traditional ROMs have achieved remarkable success in low-dimensional, smooth parameter regimes with structured problems. However, they face fundamental barriers for (1) high-dimensional parameter spaces ($d > 50$), (2) geometry-parameterized problems, (3) nonlinear/non-smooth parameter dependence, and (4) multi-scale phenomena requiring resolution adaptivity.

Machine learning approaches, particularly physics-informed neural networks and neural operators, promise to overcome these limitations through flexible nonlinear representations, mesh-free formulations, and data-driven adaptivity. The transition from linear subspace methods to nonlinear operator learning represents a paradigm shift in computational science.

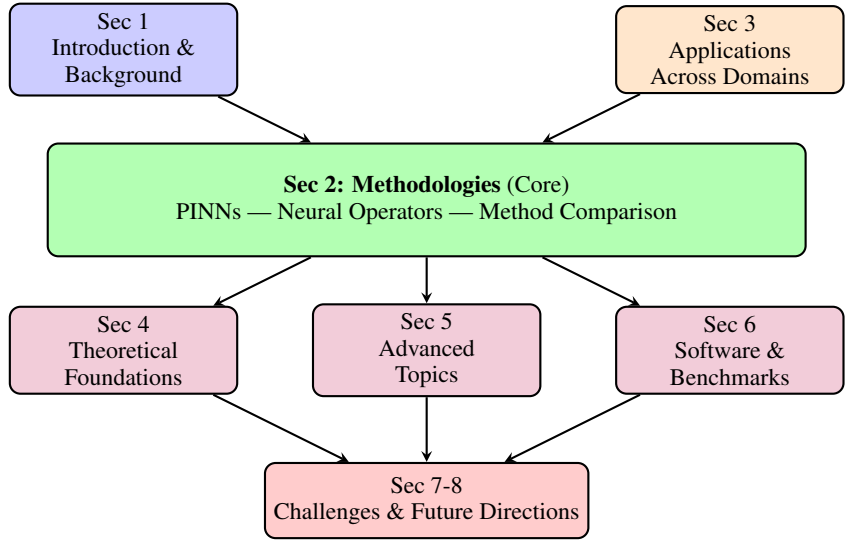
\begin{figure}[htbp]
\centering
\begin{tikzpicture}[
    section/.style={rectangle, draw, thick, rounded corners, align=center, 
                    minimum width=3cm, minimum height=1cm, font=\small},
    arrow/.style={-stealth, thick}
]

\node[font=\large\bfseries] at (6, 8) {Survey Organization};

\node[section, fill=blue!20] (s1) at (2, 6.5) {Sec 1\\Introduction \&\\Background};

\node[section, fill=green!30, minimum width=10cm, minimum height=1.5cm] (s3) at (6, 4.5) 
    {\textbf{Sec 2: Methodologies} (Core)\\PINNs | Neural Operators | Method Comparison};

\node[section, fill=orange!20] (s4) at (10, 6.5) {Sec 3\\Applications\\Across Domains};

\node[section, fill=purple!20] (s5) at (2, 2.5) {Sec 4\\Theoretical\\Foundations};
\node[section, fill=purple!20] (s6) at (6, 2.5) {Sec 5\\Advanced\\Topics};
\node[section, fill=purple!20] (s7) at (10, 2.5) {Sec 6\\Software \&\\Benchmarks};

\node[section, fill=red!20] (s8) at (6, 0.5) {Sec 7-8\\Challenges \& Future Directions};

\draw[arrow] (s1) -- (s3);
\draw[arrow] (s4) -- (s3);
\draw[arrow] (s3) -- (s5);
\draw[arrow] (s3) -- (s6);
\draw[arrow] (s3) -- (s7);
\draw[arrow] (s5) -- (s8);
\draw[arrow] (s6) -- (s8);
\draw[arrow] (s7) -- (s8);

\end{tikzpicture}
\caption{Survey organization and reading roadmap.}
\label{fig:survey_structure}
\end{figure}

Having established the mathematical foundations and traditional baseline methods, we now turn to the main focus of this survey: how modern machine learning approaches revolutionize parametric PDE solving. The survey organization is shown in Figure~\ref{fig:survey_structure}.

\section{Methodologies: Physics-Informed Neural Networks and Neural Operators}
\label{sec:methodologies}

\begin{figure}[htbp]
\centering
\begin{tikzpicture}[
    component/.style={rectangle, draw, thick, rounded corners, align=center, font=\footnotesize},
    label/.style={font=\small\bfseries}
]

\node[component, fill=yellow!30, minimum width=4cm, minimum height=1.2cm] (center) at (6, 4) 
    {\Large\textbf{Parametric PDEs}\\$\mathcal{L}(u;\mu)=f$};


\node[label, text=blue] at (2, 7) {Traditional Methods};
\node[component, fill=blue!10] at (1.5, 6) {FEM/FVM};
\node[component, fill=blue!10] at (3.5, 6) {Reduced Basis};
\node[component, fill=blue!10] at (2.5, 5.2) {POD-Galerkin};

\node[label, text=green!50!black] at (10, 7) {\textbf{Neural Methods}};
\node[component, fill=green!20] at (9, 6.3) {PINNs};
\node[component, fill=green!20] at (11, 6.3) {DeepONet};
\node[component, fill=green!20] at (9, 5.5) {FNO};
\node[component, fill=green!20] at (11, 5.5) {GNO};

\node[label, text=orange] at (2, 1.2) {Applications};
\node[component, fill=orange!10, text width=1.8cm] at (0.7, 2) {Fluid\\Dynamics};
\node[component, fill=orange!10, text width=1.8cm] at (3, 2) {Structural\\Mechanics};
\node[component, fill=orange!10, text width=1.8cm] at (0.7, 0.5) {Heat\\Transfer};
\node[component, fill=orange!10, text width=1.8cm] at (3, 0.5) {Electro-\\magnetics};

\node[label, text=red] at (10, 1.2) {Key Challenges};
\node[component, fill=red!10, text width=2cm] at (9, 2) {High-dim\\Parameters};
\node[component, fill=red!10, text width=2cm] at (11.5, 2) {Out-of-\\distribution};
\node[component, fill=red!10, text width=2cm] at (9, 0.5) {Uncertainty\\Quantification};
\node[component, fill=red!10, text width=2cm] at (11.5, 0.5) {Complex\\Geometry};

\draw[-stealth, thick, blue,line width=2pt] (2.5, 4.8) -- (center);
\draw[-stealth, thick, green!50!black, line width=2pt] (10, 4.8) -- (center);
\draw[-stealth, thick, orange,line width=2pt] (center) -- (2, 2.8);
\draw[-stealth, thick, red,line width=2pt] (center) -- (10, 2.8);

\end{tikzpicture}
\caption{Research landscape of parametric PDE solving methods.}
\label{fig:research_landscape}
\end{figure}
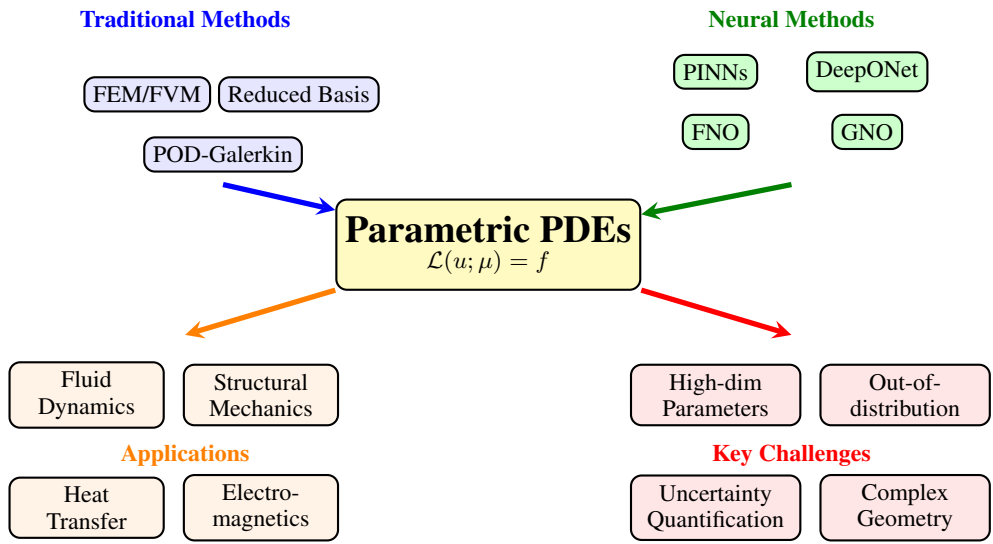

This section forms the technical core of the survey, systematically examining methods organized by their fundamental approach to parametric PDE solving. We distinguish two main paradigms: (1) \textit{physics-informed neural networks} (PINNs), which embed governing equations as soft constraints and typically require retraining for each parameter region, and (2) \textit{neural operators}, which learn mappings between infinite-dimensional function spaces and naturally generalize across parameter spaces through a single training phase. This survey covers traditional and neural methods (see Figure~\ref{fig:research_landscape}).

\subsection{Physics-Informed Neural Networks (PINNs)}
\label{sec:pinns}

\subsubsection{Foundational Framework}

Physics-informed neural networks, pioneered by Raissi et al. \citep{Raissi2019} in their seminal 2019 work, represent a breakthrough in integrating physical laws with data-driven learning. The core innovation lies in encoding PDE residuals directly into the loss function, enabling networks to discover solutions satisfying both data constraints and governing equations.

For a parametric PDE of the form \eqref{eq:parametric_pde}, a PINN approximates the solution as:
\begin{equation}
u_\theta(x,t,\mu) \approx u(x,t;\mu)
\end{equation}
where $\theta$ denotes the neural network parameters (weights and biases). The network is trained by minimizing a composite loss function:
\begin{equation}
\mathcal{L}_{\text{total}} = \mathcal{L}_{\text{PDE}} + \lambda_{\text{BC}} \mathcal{L}_{\text{BC}} + \lambda_{\text{IC}} \mathcal{L}_{\text{IC}} + \lambda_{\text{data}} \mathcal{L}_{\text{data}}
\label{eq:pinn_loss}
\end{equation}
where each term enforces different physical and data constraints:

\textbf{Physics Residual Loss:} Measures PDE satisfaction at collocation points:
\begin{equation}
\mathcal{L}_{\text{PDE}} = \frac{1}{N_{\text{PDE}}} \sum_{i=1}^{N_{\text{PDE}}} \left|\mathcal{L}(u_\theta(x_i,t_i,\mu_i); \mu_i) - f(x_i,t_i;\mu_i)\right|^2
\end{equation}
where $\{(x_i,t_i,\mu_i)\}$ are randomly sampled collocation points in the spatiotemporal-parameter domain.

\textbf{Boundary Condition Loss:} Enforces boundary constraints:
\begin{equation}
\mathcal{L}_{\text{BC}} = \frac{1}{N_{\text{BC}}} \sum_{j=1}^{N_{\text{BC}}} \left|\mathcal{B}(u_\theta(x_j,t_j,\mu_j); \mu_j) - g(x_j,t_j;\mu_j)\right|^2
\end{equation}

\textbf{Initial Condition Loss:} For time-dependent problems:
\begin{equation}
\mathcal{L}_{\text{IC}} = \frac{1}{N_{\text{IC}}} \sum_{k=1}^{N_{\text{IC}}} \left|u_\theta(x_k,0,\mu_k) - u_0(x_k;\mu_k)\right|^2
\end{equation}

\textbf{Data Fidelity Loss:} Matches available observations:
\begin{equation}
\mathcal{L}_{\text{data}} = \frac{1}{N_{\text{data}}} \sum_{\ell=1}^{N_{\text{data}}} \left|u_\theta(x_\ell,t_\ell,\mu_\ell) - u_{\ell}^{\text{obs}}\right|^2
\end{equation}

The automatic differentiation capability of modern deep learning frameworks enables efficient computation of PDE residuals through repeated backpropagation.

\textbf{Parametric Extension Strategies:}

For parametric problems, three main approaches incorporate parameters into the PINN architecture:

\textit{Strategy 1: Parameter as Additional Input} (Most Common)
\begin{equation}
u_\theta(x,t,\mu) \leftarrow \text{NN}_\theta(\text{concat}([x,t,\mu]))
\end{equation}
Parameters are concatenated with spatiotemporal coordinates as network inputs. This direct approach allows a single network to handle multiple parameter values simultaneously.

\begin{algorithm}[h]
\caption{Parametric PINN Training}
\label{alg:pinn}
\begin{algorithmic}[1]
\STATE \textbf{Input:} PDE operator $\mathcal{L}$, parameter space $\mathcal{P}$, domain $\Omega \times [0,T]$
\STATE \textbf{Initialize:} Neural network $u_\theta$ with random weights
\FOR{epoch $= 1$ to $N_{\text{epochs}}$}
    \STATE Sample collocation points: $\{(x_i,t_i,\mu_i)\}_{i=1}^{N_{\text{PDE}}}$
    \STATE Sample boundary points: $\{(x_j,t_j,\mu_j)\}_{j=1}^{N_{\text{BC}}}$
    \STATE Sample initial points: $\{(x_k,0,\mu_k)\}_{k=1}^{N_{\text{IC}}}$
    \FOR{each point $(x,t,\mu)$}
        \STATE Compute $u = u_\theta(x,t,\mu)$ via forward pass
        \STATE Compute $\frac{\partial u}{\partial t}, \nabla u, \nabla^2 u$ via automatic differentiation
        \STATE Evaluate residual: $r = \mathcal{L}(u;\mu) - f(x,t;\mu)$
    \ENDFOR
    \STATE Compute total loss: $\mathcal{L}_{\text{total}}$ via Eq. \eqref{eq:pinn_loss}
    \STATE Update $\theta$ via gradient descent: $\theta \leftarrow \theta - \alpha \nabla_\theta \mathcal{L}_{\text{total}}$
\ENDFOR
\STATE \textbf{Return:} Trained network $u_\theta$
\end{algorithmic}
\end{algorithm}

\textit{Strategy 2: Multi-Task Learning Framework}
\begin{equation}
u_\theta(x,t,\mu) = h_{\theta_{\text{shared}}}(x,t) + \sum_{k=1}^K \alpha_k(\mu;\theta_{\text{task}}) \psi_k(x,t;\theta_{\text{task}})
\end{equation}
A shared feature extractor learns common structures, while parameter-specific heads capture unique behaviors. This architecture exploits commonalities across parameter values.

\textit{Strategy 3: Conditional Neural Networks}
\begin{equation}
u_\theta(x,t,\mu) = \text{NN}_{\theta(\mu)}(x,t)
\end{equation}
where $\theta(\mu) = g_\phi(\mu)$ is a hypernetwork that generates weights conditioned on $\mu$. This approach provides maximum flexibility but increases training complexity.

\textbf{Example: Parametric Burgers Equation}

To illustrate, consider the viscous Burgers equation:
\begin{equation}
\frac{\partial u}{\partial t} + u\frac{\partial u}{\partial x} = \nu \frac{\partial^2 u}{\partial x^2}, \quad x \in [0,2\pi], \quad t \in [0,1]
\end{equation}
with viscosity parameter $\nu \in [0.01, 0.1]$ and initial condition $u(x,0) = -\sin(x)$.

The PINN network takes inputs $(x,t,\nu) \in [0,2\pi] \times [0,1] \times [0.01,0.1]$ and outputs $u_\theta(x,t,\nu)$. The physics loss becomes:
\begin{equation}
\mathcal{L}_{\text{PDE}} = \mathbb{E}\left[\left(\frac{\partial u_\theta}{\partial t} + u_\theta \frac{\partial u_\theta}{\partial x} - \nu \frac{\partial^2 u_\theta}{\partial x^2}\right)^2\right]
\end{equation}
where derivatives are computed via automatic differentiation and expectation is approximated via Monte Carlo sampling over $(x,t,\nu)$.

\subsubsection{Parametric Enhancement Techniques}

Building on the foundational framework, researchers have developed numerous techniques to improve PINN performance for parametric problems. We examine the most impactful developments:

\textbf{1. Inverse Problems and Parameter Identification}

The pioneering work of Raissi et al. \citep{Raissi2019}, building on foundational deep learning approaches for PDEs \citep{Han2018, Sirignano2018, Berg2018, Weinan2019}, demonstrated PINNs' remarkable capability for inverse problems: inferring unknown parameters from sparse observational data. Complementary methodological developments \citep{Raissi2018, Cuomo2022, Lu2022} have further established PINNs as powerful tools for parameter identification. This capability is particularly valuable when parameters represent hidden physical properties.

\textit{Framework:} Parameters $\mu$ become learnable quantities alongside network weights $\theta$. The modified loss function:
\begin{equation}
\mathcal{L}_{\text{total}}(\theta,\mu) = \mathcal{L}_{\text{PDE}}(\theta,\mu) + \mathcal{L}_{\text{data}}(\theta,\mu) + \mathcal{R}(\mu)
\end{equation}
where $\mathcal{R}(\mu)$ is an optional regularization term encoding prior knowledge about parameter values.

\textit{Case Study: Hidden Fluid Mechanics} \citep{Raissi2020Science}: Raissi et al. demonstrated parameter identification for the Navier-Stokes equations from scattered velocity measurements. Given noisy observations of the velocity field, the method simultaneously reconstructs the full flow field and infers the Reynolds number, achieving relative errors below 1\% with measurements at only 0.1\% of the domain points.

The key advantage: PINNs leverage physics to interpolate between sparse measurements, dramatically reducing data requirements compared to purely data-driven methods.

\textbf{2. Adaptive Sampling Strategies}

Uniform random sampling of collocation points often yields inefficient training, particularly in parametric settings where solution complexity varies across parameter space. Adaptive sampling concentrates points in regions of high PDE residual or parameter sensitivity.

\textit{Residual-Based Adaptive Refinement (RAR):} At each training iteration, compute residuals at all collocation points and preferentially sample from high-residual regions \citep{Daw2022, Gao2023, Lu2021Adaptive, Wu2020}:
\begin{equation}
p(x,t,\mu) \propto \left|\mathcal{L}(u_\theta(x,t,\mu);\mu) - f(x,t;\mu)\right|
\end{equation}

\textit{Active Learning for Parameter Space:} For parametric problems, identify parameter values where the current model performs poorly and add training samples at those parameters. This mirrors the greedy algorithms of reduced basis methods but operates in a data-driven manner.

Recent work by Wu et al. \citep{Wu2023} demonstrated 50-70\% reductions in required training data through parameter space active learning \citep{Lookman2023, Wu2020} for Reynolds number-dependent flows.

Advanced training methods have improved PINN performance. Wang et al. \citep{Wang2025Gradient} introduced gradient alignment from a second-order optimization perspective, dynamically adjusting loss weights based on gradient alignment to achieve 30-50\% training time reductions. De Ryck et al. \citep{DeRyck2024Operator} provided an operator preconditioning perspective on physics-informed training. Hwang and Lim \citep{Hwang2024DualCone} developed dual cone gradient descent for handling conflicting gradients.

Hao et al. \citep{Hao2024PINNacle} released PINNacle at NeurIPS 2024, providing standardized PINN evaluation across 15 canonical PDEs with 20+ method variants. Miyagawa and Yokota \citep{Miyagawa2024Functional} extended PINN theory to functional differential equations with convergence guarantees.

\textbf{3. Multi-Task and Transfer Learning}

Solving parametric PDEs inherently involves multiple related tasks (one per parameter value). Multi-task learning exploits this structure:

\textit{Shared Representation Learning:} Train a single network on multiple parameter values simultaneously, encouraging the network to learn parameter-invariant features while maintaining parameter-specific outputs:
\begin{equation}
\mathcal{L}_{\text{MT}} = \sum_{i=1}^{N_{\text{tasks}}} w_i \mathcal{L}_{\text{total}}(\mu_i)
\end{equation}
where $w_i$ are task-specific weights.

\textit{Transfer Learning:} Pre-train on easy-to-solve parameter values, then fine-tune for challenging regions. For instance, train first on high-viscosity (smooth) solutions, then transfer to low-viscosity (sharper gradients) cases.

Desai et al. \citep{Desai2021}, along with comparative studies \citep{Lu2022, Geneva2022}, showed that transfer learning reduces training iterations by up to 80\% when adapting PINNs across Reynolds numbers.

\subsubsection{Recent Advances in PINN Architecture and Training}

Recent developments have significantly improved PINN training efficiency, stability, and scalability through novel architectural designs and training strategies.

\paragraph{Ill-Conditioning Analysis and Solutions}
\citet{Cao2025IllConditioning} established a fundamental connection between PINN ill-conditioning and the Jacobian matrix condition number of the PDE system. By constructing controlled systems that adjust the condition number while preserving solutions, they demonstrated that reducing the Jacobian condition number leads to faster convergence and higher accuracy. This breakthrough enabled the first successful PINN simulation of three-dimensional flow around the M6 wing at Reynolds number 5,000, representing a significant milestone for industrial-complexity problems. \citet{Zhang2025P2PINN} built upon their foundation, applying it to the parametric domain.

\paragraph{Separable Architectures}
To mitigate the curse of dimensionality (CoD) in multi-dimensional PDEs, several separable architectures have been proposed. These methods primarily focus on decomposing the high-dimensional problem. \citet{Cho2023Separable}, for instance, introduced Separable Physics-Informed Neural Networks (SPINN), which operates on a per-axis basis to dramatically reduce network forward passes and memory overhead, enabling training with over $10^7$ collocation points. Similarly, \citet{zhang2025dfs} proposed the Dynamic Feature Separation PINN (DFS-PINN), which employs an innovative input-decoupling and dynamic interaction mechanism to achieve significant computational savings. While these methods leverage separation to tackle dimensionality, the concept has also been proven effective against spectral bias. \citet{xiong32025separated} proposed the Separated-Variable Spectral Neural Network (SV-SNN), which decomposes multivariate functions into products of univariate functions integrated with adaptive Fourier features, specifically to capture high-frequency components effectively.

\paragraph{Kolmogorov-Arnold Networks for PDEs}
A paradigm shift in neural architecture comes from Kolmogorov-Arnold Networks (KAN), which replace fixed activation functions with learnable spline-based functions. \citet{Wang2024KINN} proposed Kolmogorov-Arnold-Informed Neural Networks (KINN), systematically comparing KAN and MLP across various PDE formulations including strong form, energy form, and inverse form. KINN demonstrates significant advantages in accuracy and convergence for multi-scale, singularity, stress concentration, and nonlinear hyperelasticity problems, offering better interpretability with fewer parameters.

Building on this foundation, \citet{Jacob2024SPIKANs} introduced Separable Physics-Informed Kolmogorov-Arnold Networks (SPIKANs), applying the separation of variables principle to KANs. By decomposing problems such that each dimension is handled by an individual KAN, SPIKANs drastically reduce computational complexity without sacrificing accuracy, particularly effective for higher-dimensional PDEs where collocation points grow exponentially with dimensionality.

Most recently, \citet{Zhang2025LegendKINN} proposed Legend-KINN, integrating Legendre orthogonal polynomials into the KAN architecture combined with pseudo-time stepping in backpropagation. Legend-KINN achieves 1--3 orders of magnitude faster convergence than both KAN and MLP under identical parameter settings. Further exploring this direction, \citet{xiong22025j} developed J-PIKAN, a framework based on Jacobi orthogonal polynomials.

\subsubsection{Advantages and Limitations: A Critical Analysis}

\textbf{Strengths of PINNs for Parametric Problems:}

\textit{1. Mesh-Free Formulation:} PINNs avoid mesh generation, making them naturally suited for complex and parameterized geometries. This is particularly valuable when geometry itself is a parameter.

\textit{2. Data Efficiency Through Physics:} By encoding governing equations, PINNs can learn from sparse data—often orders of magnitude less than purely data-driven methods. This is crucial in scenarios where high-fidelity simulations are expensive.

\textit{3. Inverse Problem Capability:} The ability to treat parameters as learnable variables enables powerful parameter identification frameworks, with direct applications in calibration and data assimilation.

\textit{4. Flexible Uncertainty Quantification:} Bayesian extensions like B-PINNs \citep{Yang2021} and related uncertainty quantification approaches \citep{Yang2020Bayesian, Psaros2023, Lakshminarayanan2017, Tripathy2018} naturally quantify parameter and prediction uncertainties without requiring multiple forward solves.

\textit{5. Multi-Physics Integration:} PINNs can naturally couple multiple physics through combined loss terms, useful for multi-scale parametric problems.

\textbf{Fundamental Limitations:}

\textit{1. Parameter Space Generalization:} The most critical weakness for parametric problems: standard PINNs typically require retraining for each new parameter region. While a single network can handle modest parameter ranges during training, extrapolation to out-of-distribution parameters often fails catastrophically. A network trained on $\nu \in [0.01, 0.05]$ may produce nonsensical results for $\nu = 0.1$.

\textit{2. Training Instability:} Balancing multiple loss terms in Eq. \eqref{eq:pinn_loss} remains challenging. Stiff PDEs, multi-scale problems, and high Reynolds number flows often exhibit training instabilities, with loss components competing rather than cooperating \citep{Krishnapriyan2021, Wang2022}.

\textit{3. Computational Cost:} Each training step requires evaluating PDE residuals at numerous collocation points, involving expensive automatic differentiation operations. For high-dimensional problems, training can take days to weeks on GPUs—comparable to or exceeding traditional solver costs for single parameter values.

\textit{4. Convergence Guarantees:} Rigorous convergence theory remains incomplete. While universal approximation theorems suggest that neural networks can approximate PDE solutions arbitrarily well in principle, practical convergence rates and required network sizes are problem-dependent and often unpredictable.

\textit{5. Spectral Bias:} Neural networks exhibit spectral bias toward learning low-frequency components of solutions \citep{Rahaman2019}. For parametric problems with high-frequency features (shocks, boundary layers) that vary with parameters, this bias necessitates very deep networks or specialized architectures.

\textbf{When to Use PINNs for Parametric Problems:}

PINNs excel in specific scenarios where their unique capabilities align with problem requirements. They are particularly effective for few-shot or single-query problems when solutions are needed for only a handful of parameter values, making the training cost per parameter acceptable. Their inverse problem capability makes them invaluable when parameters must be inferred from sparse observational data, leveraging physics to interpolate between measurements. In sparse data regimes where high-fidelity training data is scarce but governing physics is well-understood, PINNs can extract maximum information by encoding physical laws as constraints. Additionally, their mesh-free nature provides advantages for complex or parameterized geometries where traditional mesh generation becomes prohibitively expensive.

For multi-query parametric scenarios requiring rapid solution evaluation across broad parameter spaces, neural operators (discussed next) offer superior performance through their inherent ability to generalize across parameter distributions in a single training phase.

\textbf{Empirical Lesson:} ``PINNs pioneered physics-informed learning but face challenges in parameter space generalization. Each new parameter region often requires retraining, limiting efficiency for parametric studies. This motivates the neural operator paradigm."

\subsection{Neural Operators: Operator Learning for Parameter Space Generalization}
\label{sec:neural_operators}

In contrast to PINNs, which approximate solutions for specific parameter values, neural operators learn mappings between infinite-dimensional function spaces. This paradigm shift enables unprecedented parameter space generalization: a single trained neural operator can evaluate solutions across entire parameter distributions without retraining.

\subsubsection{Mathematical Foundations of Operator Learning}

The theoretical foundation rests on extending universal approximation from functions to operators.

\textbf{Classical Setting:} Traditional neural networks approximate mappings $f: \RR^n \to \RR^m$ between finite-dimensional spaces.

\textbf{Operator Setting:} Neural operators approximate mappings $\mathcal{G}: \mathcal{A} \to \mathcal{U}$ between infinite-dimensional function spaces, where:
$\mathcal{A}$ is the input function space (e.g., space of initial conditions, coefficient functions, or parameter-dependent forcing), and $\mathcal{U}$ is the output solution space

For parametric PDEs, the solution operator $\mathcal{G}$ defined in Eq. \eqref{eq:solution_operator} maps parameters and input functions to solutions. Neural operators aim to learn $\mathcal{G}_\theta \approx \mathcal{G}$.

\textbf{Universal Approximation for Operators:}

Chen and Chen \citep{Chen1995} established that neural networks can approximate continuous operators arbitrarily well—a result generalized by Kovachki et al. \citep{Kovachki2021} for modern neural operator architectures.

\begin{theorem}[Neural Operator Universal Approximation (Informal)]
Let $\mathcal{G}: \mathcal{A} \to \mathcal{U}$ be a continuous operator between function spaces. For any $\epsilon > 0$, there exists a neural operator architecture and parameters $\theta$ such that:
\begin{equation}
\sup_{a \in \mathcal{A}} \norm{\mathcal{G}(a) - \mathcal{G}_\theta(a)}_{\mathcal{U}} < \epsilon
\end{equation}
\end{theorem}

This theoretical foundation justifies the neural operator approach: in principle, a sufficiently large network can learn the complete parameter-to-solution map.

\textbf{Discretization Invariance:}

A crucial advantage of neural operator formulations is discretization invariance: the operator $\mathcal{G}_\theta$ operates on continuous functions, not discrete grid representations. This means a neural operator trained on coarse-resolution data can evaluate solutions on fine-resolution grids—a capability called zero-shot super-resolution that has no traditional numerical methods analogue.

Formally, if $u_h^{\text{coarse}}$ and $u_h^{\text{fine}}$ represent discretizations of the same continuous solution $u$:
\begin{equation}
\mathcal{G}_\theta(a) \text{ can be evaluated at any discretization}
\end{equation}
This property stems from operating in function space rather than grid space.

\subsubsection{DeepONet: Deep Operator Networks}
\label{sec:deeponet}

DeepONet, introduced by Lu et al. \citep{Lu2021} in \textit{Nature Machine Intelligence}, represents the first practical implementation of operator learning achieving widespread success.

\textbf{Architecture: Branch-Trunk Decomposition}

The key innovation is decomposing the operator evaluation as:
\begin{equation}
\mathcal{G}_\theta(a)(y) = \sum_{k=1}^p b_k(a) \cdot t_k(y)
\label{eq:deeponet}
\end{equation}
where:
$a$ is an input function (e.g., initial condition, coefficient function, or parameter encoding) Moreover, $y$ is the evaluation location (spatial-temporal coordinate) Additionally, $b_k: \mathcal{A} \to \RR$ are \textit{branch networks} that encode the input function Furthermore, $t_k: \Omega \times [0,T] \to \RR$ are \textit{trunk networks} that encode the evaluation location Also, $p$ is the latent dimension (typically $p \sim 100-1000$)

The branch network takes as input a discretized representation of the input function $a$ evaluated at sensor locations $\{a(x_i)\}_{i=1}^m$:
\begin{equation}
\mathbf{b} = [b_1(a), \ldots, b_p(a)] = \text{BranchNet}([a(x_1), \ldots, a(x_m)])
\end{equation}

The trunk network takes the query location $y$:
\begin{equation}
\mathbf{t} = [t_1(y), \ldots, t_p(y)] = \text{TrunkNet}(y)
\end{equation}

The final solution is computed via dot product:
\begin{equation}
\mathcal{G}_\theta(a)(y) = \mathbf{b}^\top \mathbf{t} = \sum_{k=1}^p b_k(a) \cdot t_k(y)
\end{equation}

\textbf{Parametric PDEs with DeepONet:}

For parametric problems, parameters $\mu$ can be incorporated in multiple ways:

\textit{Approach 1: Parameters as Branch Input}
\begin{equation}
\mathbf{b} = \text{BranchNet}([a(x_1), \ldots, a(x_m), \mu_1, \ldots, \mu_d])
\end{equation}
Parameters are appended to the function sensors in the branch network input.

\textit{Approach 2: Parameters as Trunk Input}
\begin{equation}
\mathbf{t} = \text{TrunkNet}([y, \mu])
\end{equation}
Parameters are concatenated with query coordinates.

\textit{Approach 3: Dual Encoding}
Both networks receive parameter information, enabling maximum expressivity:
\begin{align}
\mathbf{b} &= \text{BranchNet}([a(x_1), \ldots, a(x_m), \mu]) \\
\mathbf{t} &= \text{TrunkNet}([y, \mu])
\end{align}

\textbf{Physics-Informed DeepONet (PI-DeepONet)}

Wang et al. \citep{Wang2021SciAdv} extended DeepONet with physics-informed training, combining data and physical constraints. The loss function:
\begin{equation}
\mathcal{L}_{\text{PI-DeepONet}} = \mathcal{L}_{\text{data}} + \lambda_{\text{PDE}} \mathcal{L}_{\text{PDE}}
\end{equation}
where:
\begin{align}
\mathcal{L}_{\text{data}} &= \mathbb{E}_{a,y}[\norm{\mathcal{G}_\theta(a)(y) - \mathcal{G}(a)(y)}^2] \\
\mathcal{L}_{\text{PDE}} &= \mathbb{E}_{a,y}[\norm{\mathcal{L}(\mathcal{G}_\theta(a)(y); \mu) - f(y;\mu)}^2]
\end{align}

This hybrid approach offers remarkable data efficiency: PI-DeepONet can learn operators from as few as 5-10 solution snapshots by leveraging PDE residuals to regularize the learned mapping.

\textbf{Case Study: Parametric Burgers Equation}

Consider learning the solution operator for Burgers equation with parametric viscosity and initial conditions:
\begin{equation}
\frac{\partial u}{\partial t} + u\frac{\partial u}{\partial x} = \nu \frac{\partial^2 u}{\partial x^2}, \quad u(x,0) = a(x)
\end{equation}
where $\nu \in [0.01, 0.1]$ and $a(x)$ varies over a function space (e.g., random Gaussian processes).

\textit{Setup:}
Input function: $a(x)$ sampled at $m=100$ sensor locations, Parameters: viscosity $\nu$, and Output: $u(x,t)$ for any $(x,t)$

\textit{Training:}
Generate $N=1000$ training triplets $(a_i, \nu_i, u_i)$ where $u_i$ solves Burgers equation with IC $a_i$ and viscosity $\nu_i$, Branch network: MLP with input $[a_i(x_1), \ldots, a_i(x_m), \nu_i] \in \RR^{101}$, output $\mathbf{b}_i \in \RR^{100}$, Trunk network: MLP with input $(x,t) \in \RR^2$, output $\mathbf{t} \in \RR^{100}$, and Minimize: $\mathcal{L} = \sum_{i,j} |\mathbf{b}_i^\top \mathbf{t}_j - u_i(x_j,t_j)|^2$

\textit{Results:} Once trained, the DeepONet evaluates solutions for any new $(a,\nu)$ in $\sim$1ms on GPU, achieving relative $L^2$ errors of 1-3\% across the parameter range—including parameter values never seen during training. This demonstrates true parameter space generalization.

\textbf{Variants and Extensions:}

\textit{POD-DeepONet:} Bhattacharya et al. \citep{Bhattacharya2021} combined proper orthogonal decomposition with DeepONet. The branch network predicts POD coefficients, and the trunk network reconstructs the solution from basis functions. This reduces output dimensionality and improves data efficiency.

\textit{MIONet (Multiple-Input Operator Network):} Extends DeepONet to multiple input functions, enabling multi-physics or multi-fidelity \citep{Penwarden2023, Howard2023, Peherstorfer2018} learning \citep{Jin2022}.

\textit{Fourier-DeepONet:} Incorporates Fourier feature embeddings in the trunk network to better capture high-frequency solution components \citep{Li2022}.

\textbf{Advanced DeepONet Variants:}

Recent developments have transformed DeepONet methodology. Mandl et al. \citep{Mandl2024Separable} introduced Separable Physics-Informed DeepONet, addressing the curse of dimensionality through separable architecture combined with forward-mode automatic differentiation. This achieved linear scaling with dimensionality—for a 4D heat equation, training time reduced from 289 hours to 2.5 hours, a remarkable 100$\times$ speedup. Zhou et al. \citep{Zhang2024PAR} developed PAR-DeepONet with physical adaptive refinement, dynamically adjusting sampling based on PDE residuals and achieving up to 71.3\% accuracy improvements.

A breakthrough came from Jiao et al. \citep{Nature2025OneShotLearning} in Nature Communications, demonstrating one-shot operator learning from single solution trajectories using self-supervised learning and meta-learning, achieving 5-10\% relative errors with just one example—previously requiring hundreds. This is revolutionary for expensive simulations. Yang \citep{Yang2025DeepONet} provided comprehensive generalization bounds for DeepONet with physics-informed training, deriving error bounds in terms of Rademacher complexity. Li et al. \citep{Li2025Architectural} analyzed the trunk-branch architecture and proposed extensions allowing nonlinear interference for improved performance on nonlinear parametric PDEs. Zhong et al. \citep{Zhong2024PICDeepONet} introduced physics-informed compositional DeepONet for complex multi-physics problems with hierarchical decomposition.

\subsubsection{Fourier Neural Operator (FNO)}
\label{sec:fno}

The Fourier Neural Operator, introduced by Li et al. \citep{Li2021ICLR} at ICLR 2021, represents a breakthrough in operator learning through spectral methods. FNO achieves remarkable computational efficiency and zero-shot super-resolution capabilities by operating in the Fourier domain, making it particularly effective for parametric PDEs on periodic domains or problems amenable to Fourier analysis.

\textbf{Motivation: Why the Frequency Domain?}

Traditional convolutional neural networks have limited receptive fields, requiring many layers to capture long-range dependencies. PDE solutions often exhibit multi-scale structures where distant spatial locations couple through the governing equations. Operating in the Fourier domain offers three key advantages:

\textit{1. Global receptive field:} Every Fourier mode couples with all spatial locations after inverse transform, enabling immediate global information propagation.

\textit{2. Efficient convolution:} The convolution theorem states that convolution in physical space is multiplication in Fourier space, enabling $\mathcal{O}(N \log N)$ operations via Fast Fourier Transform (FFT) rather than $\mathcal{O}(N^2)$.

\textit{3. Multi-scale representation:} Low and high frequencies naturally encode coarse and fine solution features, respectively, facilitating multi-scale learning.

\textbf{Architecture: Spectral Convolution Layers}

The FNO architecture consists of multiple Fourier layers. Each layer transforms the input through spectral convolution:

\begin{equation}
v_{k+1}(x) = \sigma\left(W \cdot v_k(x) + (\mathcal{K} \cdot v_k)(x)\right)
\label{eq:fno_layer}
\end{equation}

where $v_k$ is the feature field at layer $k$, $W$ is a standard pointwise linear transformation, $\mathcal{K}$ is the spectral convolution operator, and $\sigma$ is a nonlinear activation (typically GELU).

The spectral convolution is the key innovation:
\begin{equation}
(\mathcal{K} \cdot v_k)(x) = \mathcal{F}^{-1}\left(R \cdot \mathcal{F}(v_k)\right)(x)
\end{equation}
where $\mathcal{F}$ denotes the Fourier transform, $R(\omega) \in \CC^{d_v \times d_v}$ is a learnable complex-valued weight in frequency space, $\omega$ represents frequency modes, and $\mathcal{F}^{-1}$ is the inverse Fourier transform.

Crucially, only low-frequency modes are retained (typically the first $k_{\max}$ modes in each dimension), acting as a learned low-pass filter that preserves essential solution features while discarding high-frequency noise.

\textbf{Parametric FNO: Incorporating Parameters}

For parametric PDEs, three main strategies integrate parameters into FNO:

\textit{Strategy A: Parameter-Modulated Spectral Filters}
\begin{equation}
R(\omega;\mu) = \text{MLP}(\mu) \odot R_{\text{base}}(\omega)
\end{equation}
A multi-layer perceptron (MLP) generates parameter-dependent scaling factors that modulate the base spectral filters. This allows different frequencies to be emphasized or suppressed based on parameter values—for instance, low-frequency modes for high-viscosity flows.

\textit{Strategy B: Parameter as Initial Channel}
\begin{equation}
v_0(x) = [\text{Lift}(a(x)), \mu_1, \ldots, \mu_d]
\end{equation}
Parameters are broadcast across the spatial domain and concatenated as additional input channels. A lifting layer $\text{Lift}$ projects the input function $a(x)$ to higher-dimensional feature space before concatenating with parameters.

\textit{Strategy C: Multi-Task Architecture}
Multiple FNO branches are trained simultaneously, with shared low-level layers and parameter-specific high-level layers, enabling efficient multi-query training.

\textbf{Geometric FNO (Geo-FNO)}

Standard FNO assumes periodic domains and regular grids. Li et al. \citep{Li2023JMLR} extended FNO to irregular geometries through several techniques:

\textit{1. Domain mapping:} Map irregular physical domain $\Omega(\mu)$ to a regular computational domain $\hat{\Omega}$ via diffeomorphisms $\Phi_\mu: \Omega(\mu) \to \hat{\Omega}$. The PDE is then solved in $\hat{\Omega}$ where standard Fourier methods apply.

\textit{2. Non-uniform FFT:} For non-periodic or irregular domains, use type-3 non-uniform FFT (NUFFT) that handles scattered data points.

\textit{3. Geometric encoding:} Embed geometric information (e.g., distance fields, curvature) as additional input channels to inform the network about domain shape.

These advances enable FNO application to geometry-parameterized problems, such as airfoil shape optimization where the domain boundary varies with parameters.

\textbf{Physics-Informed FNO (PI-FNO)}

Wang et al. \citep{Wang2022} combined FNO with physics-informed training:
\begin{equation}
\mathcal{L}_{\text{PI-FNO}} = \mathcal{L}_{\text{data}} + \lambda_{\text{PDE}} \mathcal{L}_{\text{PDE}} + \lambda_{\text{BC}} \mathcal{L}_{\text{BC}}
\end{equation}

The PDE residual loss is computed by differentiating the FNO prediction:
\begin{equation}
\mathcal{L}_{\text{PDE}} = \mathbb{E}\left[\left|\mathcal{L}(\mathcal{G}_\theta(a)(x,t);\mu) - f(x,t;\mu)\right|^2\right]
\end{equation}

Automatic differentiation computes spatial derivatives of FNO outputs efficiently due to the spectral representation—derivatives in Fourier space are simply multiplication by $i\omega$.

\textbf{Case Study: Navier-Stokes at Varying Reynolds Numbers}

Li et al. \citep{Li2021ICLR}, building on earlier work in turbulence modeling \citep{Wang2020Turb, Kurth2023, List2023}, demonstrated FNO on 2D turbulent flows governed by the Navier-Stokes vorticity equation:
\begin{equation}
\frac{\partial \omega}{\partial t} + u \cdot \nabla \omega = \frac{1}{\text{Re}} \nabla^2 \omega + f, \quad \nabla \cdot u = 0
\end{equation}

\textit{Setup:} The experiments consider turbulent flows with Reynolds numbers ranging from 1000 to 10000. The network takes as input the initial vorticity $\omega_0(x)$ and random forcing $f(x,t)$, and outputs the vorticity field $\omega(x,t)$ at future times. Training data consists of 1000 direct numerical simulations on $256 \times 256$ grids. The FNO architecture employs 4 Fourier layers, keeping the 12 lowest modes per dimension.

\textit{Results:} The trained FNO achieves a relative $L^2$ error of 1.8\% averaged across Reynolds numbers, including unseen Re values through both interpolation and modest extrapolation. Inference time is remarkably fast at 0.005s per query on GPU compared to 5-10 minutes for DNS, yielding a speedup of approximately 60,000$\times$. The model exhibits zero-shot super-resolution capability, generalizing from training on $64 \times 64$ data to $256 \times 256$ evaluation with less than 5\% error increase. Furthermore, autoregressive rollout for $t \in [0,50]$ maintains physical properties including energy conservation and enstrophy cascade, demonstrating long-time stability.

This demonstrates FNO's capability for multi-query parametric scenarios: one training session enables rapid exploration of the Reynolds number space.

\textbf{FNO Enhancements:}

FNO has been enhanced through refinements addressing computational efficiency and accuracy. Li et al. \citep{Li2025DFNO} introduced the Decomposed Fourier Neural Operator (D-FNO) in CMAME 2025, leveraging tensor decomposition to reduce 3D complexity from $O(N^3 \log N)$ to $O(PN \log N)$, achieving 2-3$\times$ speedup over Factorized FNO while maintaining accuracy. Qin et al. \citep{Qin2024SpecBoost} developed SpecBoost-FNO addressing frequency bias through ensemble learning, achieving 50\% average accuracy improvement across benchmark problems. Kong et al. \citep{Kong2025Seismic} applied spectral-boosted FNO to 3D seismic wavefield modeling, reducing frequency bias by 40\%.

For inverse problems, Behroozi et al. \citep{Feng2025SCFNO} introduced Sensitivity-Constrained FNO (SC-FNO) at ICLR 2025, integrating sensitivity analysis into the operator framework for improved parameter inversion. Zhang et al. \citep{Zhang2025PIFNO} developed physics-informed FNO with enhanced constraint enforcement mechanisms.

Conservation laws received special attention: Cardoso-Bihlo and Bihlo \citep{CardoboBihlo2025Conservative} developed exactly conservative physics-informed neural operators in Neural Networks 2025, ensuring discrete conservation of mass, momentum, and energy through learnable adaptive correction. Liu and Tang \citep{Liu2025DiffFNO} combined diffusion models with FNO (DiffFNO) at CVPR 2025 for better uncertainty quantification and robustness, achieving 15-25\% accuracy improvements. Kalimuthu et al. \citep{Kalimuthu2025LOGLOFNO} proposed LOGLO-FNO at ICLR 2025 for efficient local-global feature learning. Liu et al. \citep{Loeffler2025CMAME} provided rigorous error analysis for FNO on parametric PDEs, establishing convergence rates.

For variable geometries, Zhong and Meidani \citep{Zhong2025PIGANO} developed Physics-Informed Geometry-Aware Neural Operator (PI-GANO) in CMAME 2025, simultaneously generalizing across PDE parameters and domain geometries using signed distance functions and parameter-geometry attention mechanisms, achieving <3\% relative errors across diverse geometric configurations without expensive FEM data generation.

\textbf{Comparative Analysis: FNO vs DeepONet}

Table \ref{tab:fno_vs_deeponet} compares the two dominant neural operator architectures for parametric PDEs.

\begin{table}[h]
\centering
\caption{Comparison of FNO and DeepONet for parametric PDE solving}
\label{tab:fno_vs_deeponet}
\small
\begin{tabular}{p{3.5cm}p{5cm}p{5cm}}
\toprule
\textbf{Aspect} & \textbf{FNO} & \textbf{DeepONet} \\
\midrule
\textbf{Core Mechanism} & Spectral convolution in Fourier space & Branch-trunk factorization \\
\midrule
\textbf{Parameter Generalization} & Strong (especially for smooth parameter dependence) & Strong (flexible parameter encoding) \\
\midrule
\textbf{Training Data} & High (1000-10000 samples) & Moderate (100-1000 samples) \\
\midrule
\textbf{Inference Speed} & Fastest ($\mathcal{O}(N \log N)$ via FFT) & Fast ($\mathcal{O}(N_{\text{query}})$) \\
\midrule
\textbf{Complex Geometry} & Moderate (requires Geo-FNO or NUFFT) & Strong (mesh-independent) \\
\midrule
\textbf{Zero-shot Super-resolution} & Native capability & Limited (requires sensor distribution design) \\
\midrule
\textbf{Physics Constraints} & Optional via PI-FNO & Natural via PI-DeepONet \\
\midrule
\textbf{Best Use Cases} & Periodic/regular domains, turbulent flows, time-dependent evolution & Irregular geometries, sparse sensors, inverse problems \\
\bottomrule
\end{tabular}
\end{table}

\textbf{Performance Insights:}
FNO excels when solutions have strong spectral content and domains are regular Moreover, DeepONet is preferable for complex geometries or when input functions are sparsely sampled Additionally, For high Reynolds number flows, FNO's spectral bias toward smooth functions can be problematic; hybrid approaches help Furthermore, Data efficiency: PI-DeepONet often requires 5-10$\times$ less data than FNO by leveraging physics Also, Computational cost: FNO training is faster per epoch but may need more epochs for convergence

\subsubsection{Graph Neural Operators and Advanced Architectures}
\label{sec:gno_advanced}

While DeepONet and FNO dominate the neural operator landscape, several emerging architectures address specific limitations or application scenarios.

\textbf{Graph Neural Operator (GNO)}

Proposed by Li et al. \citep{Li2020}, with extensions in \citep{Li2023GNOT, Hao2023GNOT2, Gao2021, Pfaff2021, Sanchez2020}, GNO represents functions on graphs rather than regular grids, naturally handling unstructured meshes arising in finite element simulations or point clouds from experimental measurements.

\textit{Key idea:} Represent the function $a(x)$ as a graph signal on vertices $\{v_i\}$: $a = [a(v_1), \ldots, a(v_N)]$. Graph convolutional layers learn the operator:
\begin{equation}
[\mathcal{G}_\theta(a)]_i = \sigma\left(\sum_{j \in \mathcal{N}(i)} W_{ij} a_j + b_i\right)
\end{equation}
where $\mathcal{N}(i)$ denotes neighbors of vertex $i$.

\textit{Parametric extension:} For geometry-parameterized problems, the graph structure itself varies with $\mu$. GNO handles this by learning message-passing rules invariant to graph topology. Parameters can modulate edge weights: $W_{ij}(\mu) = \text{MLP}(\mu, \|v_i - v_j\|, \text{edge features})$.

\textit{Applications:} Particularly effective for finite element meshes in structural mechanics, where geometry changes significantly (e.g., topology optimization, crack propagation with varying paths).

\textbf{Graph-Based Extensions:}

Graph neural operators have seen significant methodological advances for challenging geometries. Liao et al. \citep{Liao2025Curvature} developed curvature-aware graph attention at ICML 2025 for PDEs on manifolds, incorporating Riemann curvature tensors into message passing for accurate differential equation solving on curved surfaces. Lino et al. \citep{Lino2025Diffusion} introduced diffusion graph networks at ICLR 2025 for complex fluid simulations with irregular boundaries, using learned diffusion processes on graphs. Zou et al. \citep{Zou2025FDIGN} proposed finite-difference-informed graph networks in Physics of Fluids 2025 for incompressible flows on block-structured grids, elegantly combining GNN flexibility with finite-difference structure.

\textbf{Multipole Graph Neural Operator (MGNO)}

Li et al. \citep{Li2023NeurIPS} introduced MGNO to efficiently handle long-range interactions in graph-based operators. The key innovation is a multipole expansion separating near-field (local) and far-field (global) interactions:
\begin{equation}
[\mathcal{G}_\theta(a)]_i = \text{Local}(a, \mathcal{N}_{\text{near}}(i)) + \text{Global}(a, \text{multipole coefficients})
\end{equation}

This reduces computational complexity from $\mathcal{O}(N^2)$ to $\mathcal{O}(N \log N)$ for large-scale problems, enabling neural operators on million-node meshes.

\textbf{U-Net Based Operators}

Gupta and Brandstetter \citep{Gupta2022} adapted the U-Net encoder-decoder architecture for operator learning. The multi-scale nature of U-Nets—with skip connections preserving fine details—makes them effective for multi-scale parametric PDEs.

\begin{equation}
\mathcal{G}_\theta(a)(x) = \text{Decoder}_\theta(\text{Encoder}_\theta(a))(x)
\end{equation}

\textit{Parametric integration:} Parameters influence multiple scales through:
\textit{Encoder conditioning:} $\text{Encoder}_\theta(a;\mu)$ modulates downsampling based on $\mu$, \textit{Bottleneck injection:} Inject $\mu$ at the U-Net bottleneck where global information is processed, and \textit{Decoder conditioning:} Parameter-dependent upsampling for reconstruction

\textit{Advantage:} U-Net's hierarchical structure naturally handles problems where parameters control multiple length scales (e.g., turbulent flows where Re affects both large eddies and small dissipative scales).

\textbf{Transformer-Based Operators}

Recent work \citep{Cao2021, Hao2023} explores attention mechanisms for operator learning. The Transolver \citep{Wu2023Transolver, Cao2023} architecture uses:
\begin{equation}
\text{Attention}(Q,K,V) = \text{softmax}\left(\frac{QK^T}{\sqrt{d_k}}\right)V
\end{equation}
where queries $Q$, keys $K$, and values $V$ are derived from the input function and parameters.

\textit{Parameter-aware attention:} Parameters modulate attention weights:
\begin{equation}
\text{Attention}(Q,K,V;\mu) = \text{softmax}\left(\frac{QK^T + B(\mu)}{\sqrt{d_k}}\right)V
\end{equation}
where $B(\mu)$ is a parameter-dependent bias learned by an MLP.

\textit{Benefits:} Transformers capture long-range dependencies without the periodicity requirements of FNO, and their flexibility accommodates diverse parameter types (physical, geometric, boundary conditions) in a unified framework.

\textbf{Alternative Architectures:}

Emerging paradigms explore fundamentally new approaches to operator learning. Zheng et al. \citep{Zheng2024NIPS} introduced the Mamba Neural Operator at NeurIPS 2024, applying state-space models to PDEs. Mamba uses adaptive state-space matrices with alias-free design for capturing long-range spatiotemporal dependencies at reduced computational cost compared to transformers, achieving competitive performance with better scaling.

Memory-enhanced models address long-time integration challenges. Buitrago-Restrepo et al. \citep{BuitragoRestrepo2025ICLR} demonstrated explicit memory mechanisms for time-dependent PDEs at ICLR 2025. By maintaining memory buffers storing spatiotemporal patterns, they significantly reduced autoregressive errors for long-time integration (500 vs 100 stable time steps, 40\% lower error). Le Boudec et al. \citep{LeBoudec2025ICLR} and Moro et al. \citep{Moro2025ICLR} proposed novel attention mechanisms specifically designed for operator learning. Li et al. \citep{Li2025Maxup} introduced max-up training strategies that improve out-of-distribution generalization by augmenting training with worst-case parameter perturbations.

\textit{Challenges:} Quadratic complexity in the number of discretization points limits scalability; hybrid local-global attention mechanisms mitigate this.

\subsection{Method Comparison and Selection Guidelines}
\label{sec:method_comparison}

Having reviewed the major methodologies, we now provide practical guidance for selecting appropriate methods based on problem characteristics and requirements.

\subsubsection{Method Characteristics}

Different neural PDE methods have distinct characteristics suited for specific scenarios:

\textbf{Parameter Generalization:} Neural operators (DeepONet, FNO, GNO) learn mappings that generalize across parameter spaces without retraining, while PINNs typically require retraining for each parameter region.

\textbf{Data Requirements:} Physics-informed methods (PINNs, PI-DeepONet) can work with sparse or no paired training data by leveraging PDE residuals. Data-driven operators (vanilla FNO, DeepONet) require hundreds to thousands of high-fidelity solutions.

\textbf{Computational Cost:} 
\begin{itemize}
\item FNO inference: $\mathcal{O}(N \log N)$ via FFT
\item DeepONet inference: $\mathcal{O}(N_{\text{query}} \cdot d_{\text{latent}})$
\item PINN inference: $\mathcal{O}(N_{\text{collocation}} \cdot \text{network depth})$
\end{itemize}

\textbf{Geometric Flexibility:} Graph neural operators handle arbitrary mesh topologies naturally. PINNs work well on irregular domains. FNO excels on regular/periodic domains but requires extensions (Geo-FNO) for complex geometries.

\textbf{Evaluation Criteria Definitions:}
\textit{Parameter Generalization:} Ability to accurately predict solutions for parameter values outside the training distribution \textit{Training Data:} Number of high-fidelity simulations required (Low: <100, Medium: 100-1000, High: >1000) \textit{Training Time:} Wall-clock time on modern GPUs (Short: <1 hour, Medium: 1-10 hours, Long: >10 hours) \textit{Inference:} Query time per new parameter (Slow: >1s, Fast: 0.01-1s, Very Fast: <0.01s) \textit{Complex Geometry:} Handling of irregular, parameterized, or time-varying domains \textit{High-Dim Parameters:} Performance when $d > 50$

\subsubsection{Method Selection Guide}

The following text outlines the evaluation steps for selecting an appropriate method:

\textbf{Step 1: Data Availability Assessment}

\textit{Question:} How much training data is available or can be generated?

\textbf{Abundant data (>1000 samples):} Consider pure data-driven approaches Regular domain → \textit{FNO} (fastest inference), and Irregular domain → \textit{GNO} or \textit{DeepONet} \textbf{Moderate data (100-1000 samples):} Use hybrid physics-data methods Multiple parameter queries expected → \textit{PINO} or \textit{PI-DeepONet}, and Single or few queries → \textit{PINN} with transfer learning \textbf{Scarce data (<100 samples):} Physics known → \textit{PINN} or \textit{PI-DeepONet}, and Physics unknown → Traditional methods or experimental design for more data

\textbf{Step 2: Query Pattern Analysis}

\textit{Question:} How many parameter configurations will be queried?

\textbf{Single query:} PINN or traditional solver (no need for operator learning) \textbf{Few queries (2-10):} PINN with transfer learning or multi-task PINN \textbf{Many queries (>10):} Neural operators essential Real-time requirements (ms inference) → \textit{FNO} (if geometry allows) or \textit{L-DeepONet}, and Standard requirements (sub-second) → \textit{DeepONet}, \textit{PINO}, or \textit{Geo-FNO}

\textbf{Step 3: Geometric Complexity}

\textit{Question:} What is the nature of the spatial domain?

\textbf{Regular/periodic:} FNO (optimal), \textbf{Fixed irregular:} Geo-FNO or GNO, \textbf{Parameterized geometry:} GNO or PINN, and \textbf{Time-varying geometry:} GNO with dynamic graphs or moving-mesh PINN

\textbf{Step 4: Parameter Dimensionality}

\textit{Question:} How many parameters ($d$)?

\textbf{Low ($d < 10$):} All methods applicable \textbf{Medium ($d = 10$-$50$):} Neural operators with active subspace methods \textbf{High ($d > 50$):} Dimensionality reduction (active subspaces, POD) + neural operators, L-DeepONet with latent parameter encoding, and Consider whether all parameters are truly active (sensitivity analysis)

\textbf{Step 5: Physical Constraints Importance}

\textit{Question:} Are conservation laws, symmetries, or boundary conditions critical?

\textbf{Critical (safety, physical validity):} Physics-informed methods PINN (hard-codes physics), and PI-DeepONet or PINO (soft physics constraints) \textbf{Important but flexible:} PINO with physics loss weighting \textbf{Not critical (pure prediction):} Data-driven FNO, DeepONet, or GNO

\subsubsection{Application-Specific Recommendations}

We conclude this section with concrete recommendations for common parametric PDE scenarios:

\textbf{Scenario 1: Airfoil Optimization (Reynolds Number and Shape Parameters)}

\textit{Characteristics:} $d \sim 5$-$10$ (Re + shape parameters), complex geometry, many design evaluations needed, CFD data expensive.

\textit{Recommendation:} Geo-FNO or GNO
Generate $\sim$500-1000 CFD simulations covering parameter space Moreover, Use Geo-FNO if shape parameterization is smooth (B-splines, NURBS) Additionally, Use GNO if meshes vary significantly or topology changes Furthermore, Expected speedup: 1000-10,000$\times$ vs CFD per query Also, Accuracy: 2-5\% relative error typical

\textit{Alternative:} PINO if data budget is limited to <200 samples, leveraging physics to compensate.

\textbf{Scenario 2: Parameter Identification from Sparse Measurements}

\textit{Characteristics:} Unknown material properties, few sensors, single or few parameter sets of interest, physics well-understood.

\textit{Recommendation:} PINN
Formulate as inverse problem with learnable parameters, Leverage physics constraints to interpolate between sensors, Can work with as few as 5-10 sensors, and Uncertainty quantification via B-PINN

\textit{Rationale:} Neural operators require abundant data; PINN's physics encoding compensates for data scarcity in single-query scenarios.

\textbf{Scenario 3: Uncertainty Quantification with 100+ Parameters}

\textit{Characteristics:} High-dimensional parameter space (material uncertainties, manufacturing tolerances), need for probability distributions of quantities of interest, many MC samples required.

\textit{Recommendation:} Active subspace + DeepONet or PINO
Perform sensitivity analysis to identify active subspace (typically $d_{\text{eff}} \sim 5$-$10$ even if nominal $d > 100$), Train neural operator on reduced parameter space, Use trained operator for MC sampling (millions of cheap evaluations), and Validate with few full-fidelity samples

\textit{Alternative:} Sparse grid collocation if parameter dependence is smooth and $d < 30$.

\textbf{Scenario 4: Medical Imaging and Patient-Specific Modeling}

\textit{Characteristics:} Each patient has unique geometry (anatomy), need for rapid diagnosis, limited data per patient, large population dataset available.

\textit{Recommendation:} DIMON (Diffeomorphic Mapping Operator Network) or GNO
Train on population of anatomies, Use diffeomorphic mapping to canonical domain (DIMON), Or directly use graph neural operators on patient meshes (GNO), and Enables real-time patient-specific simulations

\textit{Example:} Yin et al. \citep{Yin2024} demonstrated 1006 cardiac geometries with <2\% error in electrophysiology simulations.

\textbf{Key Takeaway:} Method selection critically depends on the interplay between data availability, query frequency, geometric complexity, and parameter dimensionality. Hybrid physics-data approaches (PINO, PI-DeepONet) often provide the best balance, combining the data efficiency of physics-informed methods with the parameter generalization of neural operators.

\section{Applications Across Scientific Domains}
\label{sec:applications}

Having established the methodological foundations, we now examine how physics-informed neural networks and neural operators are transforming computational science across major application domains. This section demonstrates that the promise of rapid parameter space exploration translates to practical impact in fluid dynamics, structural mechanics, heat transfer, and electromagnetics. We emphasize parametric aspects throughout: how parameters enter each problem class, what computational challenges they pose, and how neural methods achieve breakthroughs.

\subsection{Fluid Dynamics}
\label{sec:app_fluids}

Fluid dynamics presents some of the most compelling applications for parametric neural PDE solvers. The governing Navier-Stokes equations involve natural parameters (Reynolds, Mach, and Péclet numbers) that control flow regimes, while engineering applications demand parameter sweeps for design optimization. The computational expense of traditional CFD—particularly for turbulent flows—makes neural operators especially attractive.

\subsubsection{Incompressible Navier-Stokes Equations}

The incompressible Navier-Stokes equations govern fluid motion at low Mach numbers:
\begin{align}
\frac{\partial \mathbf{u}}{\partial t} + (\mathbf{u} \cdot \nabla)\mathbf{u} &= -\nabla p + \frac{1}{\text{Re}} \nabla^2 \mathbf{u} + \mathbf{f}, \label{eq:ns_momentum} \\
\nabla \cdot \mathbf{u} &= 0, \label{eq:ns_continuity}
\end{align}
where $\mathbf{u}$ is velocity, $p$ is pressure, $\text{Re}$ is the Reynolds number, and $\mathbf{f}$ represents body forces. The Reynolds number $\text{Re} = UL/\nu$ (characteristic velocity $\times$ length / kinematic viscosity) is a fundamental parameter controlling the flow regime from laminar ($\text{Re} < 2000$) through transitional to turbulent ($\text{Re} > 4000$).

\textbf{Parametric Challenges:} As Reynolds number increases, solutions develop increasingly fine-scale structures. Turbulent flows at $\text{Re} \sim 10^6$ exhibit energy cascades spanning orders of magnitude in scale, requiring extremely high resolution ($10^9$-$10^{12}$ degrees of freedom) for direct numerical simulation. Sweeping across Reynolds numbers to understand transition or optimize designs becomes computationally prohibitive.

\textbf{Neural Operator Breakthroughs:}

\textit{1. Cylinder Flow Benchmark} \citep{Jin2021}: Jin et al. developed NSFnets (Navier-Stokes Flow nets) combining residual networks with physics constraints for flow around circular cylinders parameterized by Reynolds number and cylinder diameter.

\textit{Problem Setup:}
Parameters: $\text{Re} \in [40, 200]$, cylinder diameter $d \in [0.5, 1.5]$, Governing equations: Incompressible NS \eqref{eq:ns_momentum}-\eqref{eq:ns_continuity}, and Output: Velocity field $\mathbf{u}(x,y,t; \text{Re}, d)$ and pressure $p(x,y,t; \text{Re}, d)$

\textit{Methodology:} Physics-informed neural networks with hard enforcement of boundary conditions on cylinder surface and adaptive collocation point sampling based on residual magnitude.

\textit{Results:} Training on 200 CFD simulations covering the parameter space, NSFnets achieve relative $L^2$ error below 5\% for velocity and below 8\% for pressure. This translates to approximately 1000$\times$ speedup compared to CFD per query, reducing solution time from 15 minutes to 1 second. However, performance degrades for $\text{Re} > 200$ as the flow enters the transitional regime.

\textit{Key Insight:} The vortex shedding frequency and Strouhal number ($St = fd/U$) predictions matched experimental correlations within experimental uncertainty, demonstrating physical fidelity.

\textit{4. Advanced Fluid Applications:} The field has matured with production-grade applications. Qiu et al. \citep{Qiu2025DNS} performed direct numerical simulations of 3D two-phase flow using physics-informed neural networks with distributed parallel training on Journal of Fluid Mechanics 2025, handling millions of degrees of freedom for bubble dynamics and interfacial phenomena. Li et al. \citep{Li2024FV, Li2025FEM} developed finite volume-informed neural networks (Physics of Fluids 2024) and finite element-informed networks (JCP 2025) for unsteady incompressible flows, achieving data-free approaches that rely purely on physics residuals without any simulation data. Zou et al. \citep{Zou2025FDIGN} integrated finite-difference schemes into graph networks in Physics of Fluids 2025 for flows on block-structured grids, achieving state-of-the-art accuracy for complex industrial geometries.

\textit{2. FNO for Kolmogorov Flow} \citep{Li2021ICLR}: Li et al. applied Fourier Neural Operator to 2D forced turbulence, demonstrating unprecedented efficiency for parametric turbulent flow.

\textit{Problem Setup:} The study employs the vorticity formulation $\partial_t \omega + \mathbf{u} \cdot \nabla \omega = \text{Re}^{-1} \nabla^2 \omega + \nabla \times \mathbf{f}$ with Reynolds number $\text{Re} \in [1000, 10000]$ and forcing frequency as parameters, using a $256 \times 256$ spatial grid resolution.

\textit{Results (detailed):} Training on 1,000 DNS trajectories at various Re values, the FNO architecture with 4 Fourier layers retaining 12 lowest frequency modes achieves 1.8\% relative $L^2$ error averaged across test Re values. The model demonstrates remarkable autoregressive stability over 50 time units while preserving the physical energy spectrum. Its zero-shot super-resolution capability allows training on $64 \times 64$ grids while evaluating at $256 \times 256$ with less than 3\% error increase. The computational gain is dramatic, achieving 60,000$\times$ speedup with 0.005s inference time compared to 5-10 minutes per timestep for DNS.

\textit{Physical Validation:} The learned operator correctly captures the Kolmogorov $k^{-5/3}$ energy spectrum in the inertial range, reproduces the enstrophy cascade to small scales, and properly represents the Reynolds number dependence of integral length scales.

\textit{3. Cavity Flow with DeepONet} \citep{Dong2022}: Roohi et al. addressed lid-driven cavity flow—a canonical benchmark in CFD—using DeepONet with multi-fidelity \citep{Penwarden2023, Howard2023, Peherstorfer2018} data fusion.

\textit{Problem Setup:}
Parameters: Reynolds number $\text{Re} \in [100, 1000]$, cavity aspect ratio $L/H \in [1, 2]$, lid velocity profile $u_{\text{lid}}(x)$, and Challenge: Recirculation vortices whose number and position depend sensitively on Re

\textit{Innovation:} Combined high-fidelity CFD data (expensive, $N=50$ samples) with low-fidelity data (cheap, $N=500$ samples) through a bi-fidelity DeepONet architecture where branch networks at different fidelity levels share trunk network weights.

\textit{Results:}
Relative error: $<8\%$ using multi-fidelity \citep{Penwarden2023, Howard2023, Peherstorfer2018} approach vs $>15\%$ with high-fidelity-only training on same budget, Captured critical Re transition where secondary vortices appear, and Inference: 0.1s per parameter configuration

\textbf{Comparative Performance Summary:}

\begin{table}[h]
\centering
\caption{Performance comparison of neural methods for parametric incompressible flows}
\label{tab:fluid_performance}
\small
\begin{tabular}{p{2cm}p{2cm}p{1.5cm}p{1.5cm}p{2cm}p{1.5cm}p{2cm}}
\toprule
\textbf{Work} & \textbf{Method} & \textbf{Param Dim} & \textbf{Re Range} & \textbf{Training Data} & \textbf{Inference} & \textbf{Rel. Error} \\
\midrule
Jin (2021) & NSFnets & 2 & 40-200 & 200 CFD & $\sim$1s & $<5\%$ \\
Li (2021) & FNO & 1 & 1k-10k & 1000 DNS & 0.005s & $1.8\%$ \\
Dong (2022) & DeepONet & 3 & 100-1k & 50 HF + 500 LF & 0.1s & $<8\%$ \\
Traditional & CFD & N/A & Per Re & 1 solve & 10min-2hr & Reference \\
\bottomrule
\end{tabular}
\end{table}

\subsubsection{Compressible Flows and Shocks}

Compressible flows introduce additional parameters (Mach number, specific heat ratio) and computational challenges from discontinuities (shocks, contact discontinuities).

\textit{Parametric Dimensions:} Mach number $M = U/c$ (velocity / sound speed), ratio of specific heats $\gamma$, geometry parameters.

\textit{Neural Method Challenges:} Discontinuities violate smoothness assumptions underlying neural network approximation. Standard PINNs and neural operators struggle with Gibbs phenomena near shocks.

\textit{Recent Progress:} Several approaches address shock capturing:
\textit{Conservative formulation PINNs} \citep{Jagtap2020}: Enforce conservation laws in integral form to handle weak solutions, \textit{Shock-capturing neural operators} \citep{List2023}: Augment FNO with WENO-inspired local refinement near discontinuities, and \textit{Hyperbolic PINNs} \citep{Moseley2023}: Domain decomposition along characteristic lines

\subsubsection{Parametric Shape Optimization}

A compelling application of neural operators is aerodynamic shape optimization, where geometry itself is a high-dimensional parameter.

\textit{Problem:} Find airfoil shape $\mathcal{S}(\boldsymbol{\alpha})$ (parameterized by $\boldsymbol{\alpha} \in \mathbb{R}^d$, e.g., Bézier coefficients) that maximizes lift-to-drag ratio $L/D$ subject to flow equations.

\textit{Traditional Approach:} Adjoint-based optimization requiring $\sim$100-1000 CFD evaluations (weeks of computation).

\textit{Neural Operator Approach:} Train operator $\mathcal{G}: \boldsymbol{\alpha} \to (\mathbf{u}, p, L, D)$ on database of airfoil shapes, then use for rapid optimization.

\subsection{Solid Mechanics and Structural Optimization}
\label{sec:app_solids}

Solid mechanics problems involve material parameter identification, design optimization under varying loading, and geometry-parameterized structural analysis—all areas where parametric neural methods are making significant contributions.

\subsubsection{Linear Elasticity}

The linear elastic equations govern small-deformation structural response:
\begin{equation}
-\nabla \cdot (\mathbb{C}(\boldsymbol{\mu}) : \nabla \mathbf{u}) = \mathbf{f}, \quad \text{in } \Omega(\boldsymbol{\mu}_{\text{geom}})
\end{equation}
where $\mathbf{u}$ is displacement, $\mathbb{C}$ is the elasticity tensor (parameterized by Young's modulus $E$, Poisson ratio $\nu$), and $\Omega$ may be geometry-parameterized.

\textbf{Parametric Scenarios:} The problem accommodates three main types of parameters. Material parameters $\boldsymbol{\mu} = (E, \nu)$ represent Young's modulus and Poisson ratio for isotropic materials, extending to up to 21 independent components for anisotropic materials. Loading parameters capture force magnitude, spatial distribution, and directional variations. Geometry parameters encompass structural shape variations, hole positions, and thickness distributions.

\textbf{Representative Studies:}

\textit{1. Parameter Identification with PINNs} \citep{Haghighat2021}: Haghighat et al. developed a PINN framework for identifying spatially-varying material properties from displacement measurements.

\textit{Setup:}
Unknown: Heterogeneous Young's modulus field $E(x,y)$, Observations: Displacement measurements at 100 sensor locations (1\% of domain), and Prior: Smooth spatial variation, $E \in [50, 150]$ GPa

\textit{Approach:} Treat $E(x,y)$ as an additional neural network output. Minimize combined loss:
\begin{equation}
\mathcal{L} = \mathcal{L}_{\text{PDE}} + \lambda_{\text{data}} \mathcal{L}_{\text{data}} + \lambda_{\text{reg}} \|\nabla E\|^2
\end{equation}
where the regularization term enforces smoothness.

\textit{Results:}
Reconstructed $E(x,y)$ with $<3\%$ error from sparse data, Simultaneously obtained displacement field at arbitrary resolution, and Uncertainty quantification via Bayesian PINN variant showed credible intervals consistent with ground truth

\textit{2. Operator Learning for Parametric Structures} \citep{Goswami2022}: Goswami et al. applied DeepONet to bridge structures parameterized by loading patterns and geometric variations.

\textit{Problem:}
Input function: Distributed load $f(x)$ on bridge deck, Parameters: Span length $L \in [10, 30]$m, support stiffness $k$, and Output: Displacement field $\mathbf{u}(x; f, L, k)$ and stress tensor $\boldsymbol{\sigma}$

\textit{Results:}
Training: 800 FEM simulations with varying $(f, L, k)$, Accuracy: $<5\%$ error for displacement, $<10\%$ for stress (stress concentrations challenging), and Application: Real-time structural health monitoring—compare predicted vs measured displacements to detect damage

\subsubsection{Nonlinear Mechanics and Plasticity}

Nonlinear constitutive relations introduce severe challenges. Hyperelastic materials (rubbers, biological tissues) have strain energy functions $W(\mathbf{F}; \boldsymbol{\mu})$ dependent on material parameters $\boldsymbol{\mu}$ (e.g., Mooney-Rivlin constants). Elastoplastic materials exhibit history-dependent, irreversible deformation.

\textit{Neural Constitutive Modeling:} Recent work \citep{Masi2021, Thakolkaran2022} learns constitutive relations directly from data:
\begin{equation}
\boldsymbol{\sigma} = \mathcal{NN}_\theta(\mathbf{F}, \text{history}; \boldsymbol{\mu})
\end{equation}
ensuring thermodynamic consistency (e.g., positive dissipation) through architecture constraints.

\textit{Application - Elastomeric Materials:} Thakolkaran et al. \citep{Thakolkaran2022} trained neural operators to predict large-deformation response of elastomers across parameter space of Ogden model coefficients. Achieved $<2\%$ error in stress-strain curves for complex multiaxial loading, enabling rapid virtual testing for material design.

\subsubsection{Topology Optimization}

Zhu et al. \citep{Zhu2023PhaseField} introduced Phase-Field DeepONet using energy-based loss functions for pattern formation, enabling fast Allen-Cahn and Cahn-Hilliard simulations. Lee et al. \citep{Lee2025FEONO} developed FE Operator Networks for high-dimensional elasticity (d=50 parameters), achieving 60\% error reduction.

Topology optimization seeks the optimal material distribution $\rho(x) \in [0,1]$ (density) minimizing compliance subject to volume constraint:
\begin{equation}
\min_{\rho} \, c(\rho) = \int_\Omega f \cdot u(\rho) \, dx, \quad \text{s.t.} \int_\Omega \rho \, dx \leq V_{\text{max}}
\end{equation}
where $u(\rho)$ solves elasticity with density-dependent stiffness.

\textit{Parametric Aspect:} Loading conditions, boundary supports, and volume fractions are parameters defining different optimization scenarios.

\textit{Neural Approach:} Chandrasekhar and Suresh \citep{Sosnovik2019} \citep{Chandrasekhar2021} trained neural operators to map loading patterns to optimal topologies:
\begin{equation}
\mathcal{G}: f(x) \to \rho^*(x)
\end{equation}

\textit{Breakthrough:} After training on 10,000 topology optimization problems (each requiring iterative FEM), the neural operator:
Generates near-optimal designs in $<1$ second (vs hours for conventional optimization), Achieved 95-98\% of optimal compliance, Handles novel loading patterns via generalization, and Enables interactive design exploration

\textit{Impact:} Transforms topology optimization from overnight batch process to interactive design tool.

\subsubsection{Fracture Mechanics}

Crack propagation is highly parameter-sensitive: small changes in loading, material properties, or initial flaw geometry can dramatically alter fracture paths.

\textit{Phase-Field Fracture:} The phase-field approach introduces damage variable $d(x,t) \in [0,1]$ governed by coupled PDEs:
\begin{align}
-\nabla \cdot (\sigma(u,d)) &= 0, \\
d - \ell^2 \nabla^2 d &= \mathcal{H}(u) / G_c,
\end{align}
where $\ell$ is length scale, $G_c$ is fracture toughness, and $\mathcal{H}$ is elastic energy.

\textit{Parametric Challenge:} Parameter sweeps over $(G_c, \ell, \text{loading rate})$ require thousands of expensive phase-field simulations.

\textit{Neural Solution:} Goswami et al. \citep{Goswami2023} developed transfer learning PINNs:
Pre-train on simple fracture geometries (Mode I cracks), Fine-tune for complex scenarios (branching, multiple cracks), and Achieves 10$\times$ speedup per parameter value while maintaining crack path accuracy

\subsection{Heat Transfer and Conjugate Problems}
\label{sec:app_heat}

Heat transfer problems naturally involve parametric variations in thermal conductivity, convection coefficients, and heat source characteristics. Conjugate heat transfer coupling fluid flow with thermal conduction adds further complexity.

\subsubsection{Parametric Heat Conduction}

The heat equation with parameter-dependent thermal conductivity:
\begin{equation}
\rho c_p \frac{\partial T}{\partial t} = \nabla \cdot (k(\mathbf{x}; \boldsymbol{\mu}) \nabla T) + Q(\mathbf{x}, t; \boldsymbol{\mu})
\end{equation}
where $k$ is thermal conductivity and $Q$ is heat source.

\textbf{Industrial Application - Electronic Cooling:} Cai et al. \citep{Cai2021Heat} addressed thermal management of chip packages with varying power distributions and heat sink geometries.

\textit{Parameters:}
Heat generation map $Q(x,y)$ (spatially varying chip power), Heat sink fin spacing $s \in [1, 5]$ mm, and Convection coefficient $h \in [10, 100]$ W/(m²·K)

\textit{Neural Operator Solution:} Physics-informed DeepONet trained on 500 thermal simulations. Branch network encodes $Q(x,y)$ at sensor locations; parameters $(s,h)$ input to both networks.

\textit{Results:}
Accuracy: $<2°\text{C}$ error in maximum temperature prediction, Speedup: 5000$\times$ vs commercial thermal solver, and Design optimization: Evaluated 50,000 configurations in 10 minutes, identified optimal heat sink reducing peak temperature by 15°C

\subsubsection{Thermal Property Identification}

\textit{Inverse Problem:} Infer spatially-varying thermal conductivity $k(x,y)$ from limited temperature measurements—critical for characterizing novel materials or detecting defects.

\textit{PINN Approach:} Anantha Padmanabha et al. \citep{Anantha2021} developed Bayesian PINNs:
Observations: Temperature at 50 locations via infrared thermography, Unknown: 2D conductivity field $k(x,y)$ with suspected discontinuities (material interfaces), and Method: PINN with total variation regularization to preserve sharp interfaces

\textit{Achievement:} Reconstructed $k(x,y)$ including interfacial discontinuities with $<5\%$ error, providing uncertainty maps guiding sensor placement for reduced uncertainty.

\subsubsection{Conjugate Heat Transfer}

Conjugate problems couple fluid flow (Navier-Stokes) with heat conduction in solids, requiring matched temperature and heat flux at interfaces.

\textit{Parametric Complexity:} Both fluid Reynolds number and solid thermal conductivity ratios affect heat transfer, creating multi-parameter coupling.

\textit{State-of-the-Art:} Penwarden et al. \citep{Penwarden2023} developed multi-fidelity \citep{Penwarden2023, Howard2023, Peherstorfer2018} neural operators for conjugate heat transfer:
Low fidelity: Simplified 1D thermal resistance models (cheap), High fidelity: Coupled CFD-conduction simulations (expensive), and Neural fusion: Multi-fidelity DeepONet learns correction from low to high fidelity

\textit{Performance:} With 100 high-fidelity and 1000 low-fidelity samples, achieved $<3\%$ error in heat transfer coefficients across parameter ranges $(\text{Re}, k_{\text{ratio}})$, enabling design optimization of heat exchangers.

\subsection{Electromagnetics and Wave Propagation}
\label{sec:app_em}

Electromagnetic problems governed by Maxwell's equations involve material parameters (permittivity $\epsilon$, permeability $\mu$, conductivity $\sigma$) and geometric configurations, making them prime candidates for parametric neural methods.

\subsubsection{Parametric Maxwell's Equations}

Time-harmonic Maxwell equations \citep{Chen2020Maxwell, Wiecha2021} in frequency domain:
\begin{align}
\nabla \times \mathbf{E} &= -i\omega \mu(\mathbf{x}; \boldsymbol{\mu}) \mathbf{H}, \\
\nabla \times \mathbf{H} &= i\omega \epsilon(\mathbf{x}; \boldsymbol{\mu}) \mathbf{E} + \sigma \mathbf{E},
\end{align}
where $\omega$ is angular frequency and parameters $\boldsymbol{\mu}$ describe material distributions.

\textbf{Application - Metamaterial Design:} Chen et al. \citep{Chen2020Maxwell} used neural operators for inverse design \citep{Lu2021Adaptive, Meng2022} of electromagnetic metamaterials.

\textit{Problem:} Find spatial permittivity distribution $\epsilon(x,y)$ yielding desired scattering behavior (e.g., cloaking, focusing).

\textit{Approach:}
Forward operator: $\mathcal{G}_F: \epsilon(x,y) \to \mathbf{E}(x,y)$ (permittivity → field), and Inverse operator: $\mathcal{G}_I: \mathbf{E}_{\text{target}}(x,y) \to \epsilon(x,y)$ (desired field → design)

\textit{Training:} 20,000 random permittivity configurations, full-wave simulations for each.

\textit{Results:}
Forward prediction: $<2\%$ error in $\mathbf{E}$ field compared to finite-element solver, Inverse design: Generated metamaterial achieving 85-90\% of target performance, and Design time: Seconds vs days for topology optimization

\subsubsection{Acoustic Wave Propagation}

The acoustic wave \citep{Jin2022Acoustic} equation in heterogeneous media:
\begin{equation}
\frac{1}{c(\mathbf{x}; \boldsymbol{\mu})^2} \frac{\partial^2 p}{\partial t^2} = \nabla^2 p + s(\mathbf{x}, t)
\end{equation}
where $c$ is sound speed (material parameter) and $p$ is pressure.

\textit{Seismic Imaging Application:} Moseley et al. \citep{Moseley2023Seismic} addressed full-waveform inversion for subsurface velocity model estimation.

\textit{Parameters:} 2D velocity field $c(x,z)$ with typical dimension $\sim$10,000 (discretized).

\textit{Challenge:} High-dimensional parameter space, expensive forward wave simulations, local minima in optimization.

\textit{Neural Solution:} Learned neural operator $\mathcal{G}: c(x,z) \to \text{seismogram}$ enables:
Gradient-based inversion 1000$\times$ faster than adjoint-state methods, Better convergence by avoiding local minima through learned physics, and Uncertainty quantification via ensemble neural operators

\subsection{Cross-Domain Insights and Application Maturity}
\label{sec:app_summary}

\begin{table}[h]
\centering
\caption{Cross-domain comparison of neural methods for parametric PDEs}
\label{tab:cross_domain}
\small
\begin{tabular}{p{2.5cm}p{2.5cm}p{1.5cm}p{2.5cm}p{2.5cm}}
\toprule
\textbf{Domain} & \textbf{Common Parameters} & \textbf{Param Dim} & \textbf{Dominant Methods} & \textbf{Reported Speedup} \\
\midrule
Fluid Dynamics & Re, Mach, geometry & 1-10 & FNO, DeepONet, PINO & $10^3$-$10^5\times$ \citep{Li2021ICLR, Jin2021} \\
\midrule
Solid Mechanics & $E$, $\nu$, loading & 1-20 & PINN, DeepONet, GNO & $10^2$-$10^3\times$ \\
\midrule
Heat Transfer & $k$, $h$, sources & 1-10 & PI-DeepONet, PINN & $10^3$-$10^4\times$ \\
\midrule
Electromagnetics & $\epsilon$, $\mu$, $\omega$ & 1-5 & PINN, MaxwellNet & $10^2$-$10^3\times$ \\
\midrule
Acoustics & $c$, density & 1-100 & FWI operators & $10^2$-$10^3\times$ \\
\bottomrule
\end{tabular}
\end{table}

\textbf{Common Success Patterns:}
\textbf{Success Factors:} Neural methods achieve greatest impact in applications with well-characterized physics, where mature traditional solvers provide abundant training data for neural operator development. Multi-query scenarios such as design optimization, uncertainty quantification, and parameter studies justify the upfront neural training investment through repeated rapid evaluations. The methods particularly excel when solution manifolds exhibit smooth parameter dependence with low intrinsic dimensionality, enabling efficient learning from finite training samples.

\textbf{Remaining Barriers:} Despite these successes, significant challenges persist. High-Reynolds-number turbulent flows and chaotic dynamics pose difficulties for long-term stability in autoregressive predictions. Processes spanning more than six orders of magnitude in length or time scales present multi-scale coupling challenges that strain current architectures. Rare events including phase transitions, shock formations, and bifurcations demand specialized handling beyond standard training approaches.

\textbf{Industrial Adoption Status:}
\textbf{Production deployment:} Weather forecasting (FourCastNet \citep{Pathak2022, Kurth2023}), some engineering design optimization, \textbf{Pilot projects:} Digital twins, real-time control systems, medical imaging, and \textbf{Research stage:} Nuclear fusion, climate prediction, drug discovery

\section{Theoretical Foundations and Analysis}
\label{sec:theory}

While empirical success has driven adoption of neural methods for parametric PDEs, rigorous theoretical understanding is essential for reliability, interpretability, and principled algorithm design. This section examines the mathematical foundations underpinning physics-informed networks and neural operators, analyzing approximation capabilities, generalization behavior, and computational complexity with emphasis on parametric aspects.

\subsection{Approximation Theory for Parametric PDEs}
\label{sec:approximation}

\subsubsection{The Parametric Solution Manifold}

Understanding the structure of solution spaces is fundamental to approximation theory. For a parametric PDE, the solution manifold
\begin{equation}
\mathcal{M} = \{u(\cdot; \mu) : \mu \in \mathcal{P}\} \subset \mathcal{U}
\end{equation}
where $\mathcal{U}$ is an appropriate function space (typically Sobolev), encodes all possible solutions as parameters vary. The manifold's geometric properties—dimension, curvature, smoothness—determine approximation difficulty.

\textbf{Kolmogorov n-Width:} A classical measure of manifold complexity from reduced-order modeling theory \citep{Binev2011, Berkooz1993, Hesthaven2016, Peherstorfer2016} is the Kolmogorov $n$-width:
\begin{equation}
d_n(\mathcal{M}, \mathcal{U}) = \inf_{\mathcal{V}_n} \sup_{u \in \mathcal{M}} \inf_{v \in \mathcal{V}_n} \|u - v\|_{\mathcal{U}}
\end{equation}
where the infimum is over all $n$-dimensional subspaces $\mathcal{V}_n \subset \mathcal{U}$. This quantifies how well $\mathcal{M}$ can be approximated by $n$-dimensional linear subspaces.

For elliptic PDEs with smooth, affine parameter dependence, $d_n(\mathcal{M}) \sim e^{-\alpha n}$ (exponential decay) justifies reduced basis methods \citep{Binev2011}. However, transport-dominated or hyperbolic problems may have $d_n \sim n^{-\beta}$ (polynomial decay), requiring larger $n$ for adequate approximation.

\textbf{Implications for Neural Methods:} While Kolmogorov $n$-width characterizes linear approximation, neural networks provide nonlinear approximation. The relevant question: can neural operators achieve better decay rates through nonlinear representations?

\subsubsection{Universal Approximation Theorems for Operators}

Classical universal approximation for functions $f: \mathbb{R}^n \to \mathbb{R}^m$ states neural networks can approximate continuous functions arbitrarily well. For operators, we require extensions to infinite-dimensional spaces.

\begin{theorem}[Neural Operator Universal Approximation \citep{Kovachki2021}]
Let $\mathcal{A}$ and $\mathcal{U}$ be compact subsets of separable Banach spaces. For any continuous operator $\mathcal{G}: \mathcal{A} \to \mathcal{U}$ and $\epsilon > 0$, there exists a neural operator architecture with parameters $\theta$ such that:
\begin{equation}
\sup_{a \in \mathcal{A}} \|\mathcal{G}(a) - \mathcal{G}_\theta(a)\|_{\mathcal{U}} < \epsilon.
\end{equation}
\end{theorem}

\textit{Proof Sketch:} The proof proceeds in three steps:
\begin{enumerate}
\item \textit{Discretization:} Approximate continuous functions in $\mathcal{A}$ by finite-dimensional representations via sampling
\item \textit{Finite approximation:} Use classical universal approximation for the finite-dimensional case
\item \textit{Consistency:} Show approximation error vanishes as discretization refines
\end{enumerate}

\textbf{Parametric Extension:} For parametric operators $\mathcal{G}: \mathcal{P} \times \mathcal{A} \to \mathcal{U}$, the theorem applies with parameters entering either as finite-dimensional inputs (concatenated to function discretizations) or through conditioning mechanisms.

\textbf{Architecture-Specific Results:}

\textit{DeepONet:} Lu et al. \citep{Lu2021} proved that the branch-trunk decomposition \eqref{eq:deeponet} achieves universal approximation if:
Branch network width $p \to \infty$, and Both branch and trunk networks are universal approximators

The key insight: factorization $\sum_{k=1}^p b_k(a) t_k(y)$ can represent any operator via appropriate choices of basis functions.

\textit{FNO:} Li et al. \citep{Li2021ICLR} showed Fourier neural operators approximate operators by leveraging spectral properties. For periodic problems, the Fourier series representation provides natural expressivity. The critical assumption: solutions have sufficient spectral decay (smoothness in Fourier space).

\textbf{Approximation Rates:} While existence is guaranteed, rates remain an active research area. For smooth solutions:
\begin{equation}
\|\mathcal{G} - \mathcal{G}_\theta\|_{\text{op}} \lesssim \frac{1}{W^{\alpha/d}}
\end{equation}
where $W$ is network width, $d$ is input dimension, and $\alpha$ is smoothness. The curse of dimensionality ($1/W^{\alpha/d}$) persists theoretically \citep{Han2018, Beck2021, Khoo2021}, though empirically neural operators perform better than this bound suggests.

\subsubsection{Parametric PDE-Specific Analysis}

Recent work examines approximation rates specifically for parametric PDE solutions.

\textbf{Low-Dimensional Structure:} Bhattacharya et al. \citep{Bhattacharya2021} proved that if the solution manifold $\mathcal{M}$ has intrinsic dimension $d_{\text{eff}} \ll d$ (as measured by Kolmogorov width decay), then neural operators can achieve dimension-independent error bounds:
\begin{equation}
\|\mathcal{G} - \mathcal{G}_\theta\|_{\text{op}} \lesssim e^{-c W^{1/d_{\text{eff}}}}
\end{equation}
This explains empirical success in problems where traditional theory predicts exponential complexity in $d$.

\textbf{Regularity Propagation:} For parametric elliptic PDEs, if parameters enter affinely and solutions have bounded derivatives, then neural network approximation error inherits regularity:
\begin{equation}
\|u(\cdot; \mu) - u_\theta(\cdot; \mu)\|_{H^k} \lesssim C(\text{network size}, \text{depth}, k)
\end{equation}
where $H^k$ is the Sobolev space of order $k$.

\subsection{Generalization and Parameter Space Coverage}
\label{sec:generalization}

\subsubsection{Training and Generalization Error Decomposition}

For parametric neural methods, generalization encompasses two distinct aspects that must be considered jointly. Spatial generalization involves evaluating solutions at new spatial locations $x$ not present in the training set, while parameter generalization requires accurate predictions for parameter values $\mu$ outside the training distribution. Both capabilities are essential for practical parametric surrogate modeling.

The total error decomposes as:
\begin{equation}
\mathbb{E}_{\mu, x}[|u(x;\mu) - u_\theta(x;\mu)|^2] = \underbrace{\mathbb{E}[|u - u_\theta^*|^2]}_{\text{Approximation}} + \underbrace{\mathbb{E}[|u_\theta^* - u_\theta|^2]}_{\text{Generalization}}
\end{equation}
where $u_\theta^*$ is the best possible network in the function class.

\textbf{Parametric Complexity:} The generalization error depends on:
\textit{Parameter space coverage:} Training sample distribution in $\mathcal{P}$, \textit{Solution smoothness:} Lipschitz constant of $\mu \mapsto u(\cdot; \mu)$, and \textit{Network capacity:} Number of parameters relative to sample size

\subsubsection{Sample Complexity Bounds}

\textbf{Question:} How many training samples $N$ are needed to achieve $\epsilon$-accuracy over parameter space?

De Ryck and Mishra \citep{DeRyck2022} derived sample complexity for PINNs approximating parametric elliptic PDEs:
\begin{equation}
N \sim \frac{d \log(1/\epsilon)}{\epsilon^2} \cdot \text{poly}(\text{network width})
\end{equation}
where $d$ is parameter dimension. This exhibits logarithmic dependence on accuracy but polynomial dependence on dimension—significantly better than exponential.

For neural operators, Lanthaler et al. \citep{Lanthaler2022} proved:
\begin{equation}
N \sim d_{\mathcal{M}}^{1+\delta} \epsilon^{-2}
\end{equation}
where $d_{\mathcal{M}}$ is the effective dimension of the solution manifold. When $d_{\mathcal{M}} \ll d$ (low intrinsic dimensionality), sample requirements scale favorably.

\textbf{Empirical Observations:} Practical training often uses fewer samples than theoretical bounds suggest. Possible explanations:
Over-conservative bounds not accounting for problem structure, Physics constraints (in PINNs) reducing sample requirements, and Implicit regularization from optimization algorithms

\subsubsection{Interpolation vs. Extrapolation}

A critical distinction in parametric problems:
\textbf{Interpolation:} $\mu_{\text{test}} \in \text{convex hull}(\{\mu_{\text{train}}\})$, and \textbf{Extrapolation:} $\mu_{\text{test}} \notin \text{convex hull}(\{\mu_{\text{train}}\})$

\textbf{Empirical Findings:}
\textit{FNO:} Interpolation error $\sim$1-2\%, extrapolation error $\sim$10-20\% (Li et al. \citep{Li2021ICLR}), \textit{DeepONet:} Similar interpolation/extrapolation gap, slightly better extrapolation when physics-informed, and \textit{PINN:} Poor parametric generalization—essentially no extrapolation capability without retraining

\textbf{Theoretical Understanding:} Extrapolation fundamentally requires assumptions about solution structure beyond training data. Neural operators implicitly learn these through architecture inductive biases (e.g., FNO's Fourier basis assumes periodic/smooth solutions).

\subsubsection{Out-of-Distribution Detection}

For safety-critical applications, detecting when $\mu_{\text{test}}$ lies outside the reliable prediction region is crucial.

\textit{Approaches:}
Uncertainty quantification employs multiple complementary approaches. Bayesian neural operators compute predictive variance as an out-of-distribution indicator. Ensemble methods identify uncertain regions through high variance across ensemble members. Conformal prediction constructs prediction sets with guaranteed coverage \citep{Staber2024}, providing distribution-free uncertainty bounds.

\textit{Breakthrough - Conformal Prediction for PDEs:} Staber et al. \citep{Staber2024} developed distribution-free uncertainty quantification for neural operators:
\begin{equation}
\mathcal{C}(a, \mu) = \{u : \|u - \mathcal{G}_\theta(a, \mu)\|_{\mathcal{U}} \leq q_{\alpha}\}
\end{equation}
where $q_\alpha$ is the $(1-\alpha)$-quantile of calibration residuals. This provides rigorous $1-\alpha$ coverage guarantees without distributional assumptions—a major advance for reliability.

\subsection{Computational Complexity Analysis}
\label{sec:complexity}

\subsubsection{Training Complexity}

The computational cost structure differs fundamentally across methods:

\textbf{PINNs:}
\textit{Per-iteration cost:} $\mathcal{O}(N_{\text{collocation}} \cdot N_{\text{params}} \cdot \text{AD cost})$ where AD (automatic differentiation) for second derivatives is expensive, \textit{Total training:} $\mathcal{O}(N_{\text{iter}} \cdot N_{\text{collocation}} \cdot N_{\text{params}})$, typically $N_{\text{iter}} \sim 10^4$-$10^5$, and \textit{Bottleneck:} Computing PDE residuals via repeated differentiation

\textbf{DeepONet:}
\textit{Per-iteration:} $\mathcal{O}(N_{\text{data}} \cdot (N_{\text{branch}} + N_{\text{trunk}}))$ - standard forward pass, \textit{Total training:} $\mathcal{O}(N_{\text{iter}} \cdot N_{\text{data}})$, with $N_{\text{iter}} \sim 10^3$-$10^4$, and \textit{Advantage:} No automatic differentiation needed unless physics-informed

\textbf{FNO:}
\textit{Per-iteration:} $\mathcal{O}(N_{\text{data}} \cdot N_{\text{grid}} \log N_{\text{grid}})$ due to FFT, \textit{Total training:} Similar to DeepONet but with FFT overhead, and \textit{Scaling:} Exceptionally favorable for large spatial grids

\subsubsection{Inference Complexity}

This is where neural operators shine for parametric problems:

\begin{table}[h]
\centering
\caption{Inference complexity comparison for parametric PDEs}
\label{tab:complexity}
\small
\begin{tabular}{p{3cm}p{3.5cm}p{3.5cm}p{3cm}}
\toprule
\textbf{Method} & \textbf{Training (offline)} & \textbf{Inference (online)} & \textbf{Multi-Query Advantage} \\
\midrule
Traditional FEM & N/A & $\mathcal{O}(N_{\text{DOF}}^{2-3})$ per $\mu$ & None \\
\midrule
Reduced Basis & $\mathcal{O}(M \cdot N_{\text{DOF}}^{2-3})$ & $\mathcal{O}(N_{\text{RB}}^3) \ll N_{\text{DOF}}^3$ & High \\
\midrule
PINN & $\mathcal{O}(10^4 \cdot N_{\text{coll}})$ per $\mu$ & $\mathcal{O}(N_{\text{query}})$ & Low (retrain) \\
\midrule
DeepONet & $\mathcal{O}(10^3 \cdot N_{\text{data}})$ & $\mathcal{O}(N_{\text{query}})$ & Very High \\
\midrule
FNO & $\mathcal{O}(10^3 \cdot N_{\text{data}})$ & $\mathcal{O}(N \log N)$ & Very High \\
\bottomrule
\end{tabular}
\end{table}

\textbf{Break-Even Analysis:} Neural operators become advantageous when:
\begin{equation}
N_{\text{queries}} > \frac{C_{\text{training}}}{C_{\text{traditional}} - C_{\text{inference}}}
\end{equation}

For typical parameters:
DeepONet: Break-even at $\sim$10-50 queries, FNO: Break-even at $\sim$5-20 queries, and PINN: Rarely breaks even for parametric studies

\subsubsection{Memory Requirements}

\textbf{Storage:} Neural operators require storing network weights ($\sim$10-100 MB typically) vs. reduced basis methods storing basis functions ($\sim$GB for high-fidelity problems).

\textbf{Runtime Memory:} 
FNO: Requires full spatial field in memory ($\sim$GB for 3D problems), and DeepONet: Query-point-by-point evaluation possible, lower memory

\subsubsection{Parallelization and Hardware Efficiency}

\textbf{GPU Acceleration:} Neural operators achieve massive parallelization:
Batch processing across parameters: Evaluate 100s of $\mu$ simultaneously, Spatial parallelization: All grid points computed in parallel, and Training parallelization: Data parallel across multiple GPUs

\textbf{Specialized Hardware:} Emerging neuromorphic chips and tensor processing units offer 10-100$\times$ additional speedups for inference, potentially enabling microsecond-scale PDE solving.

\subsection{Convergence and Stability Theory}
\label{sec:convergence}

\subsubsection{Training Convergence for PINNs}

A fundamental challenge: proving that PINN training converges to PDE solutions. Recent progress:

\textbf{Neural Tangent Kernel (NTK) Analysis:} Wang et al. \citep{Wang2022NTK}, building on failure mode analyses \citep{Krishnapriyan2021, Krishnapriyan2022, Markidis2021, Fuks2020}, used NTK theory to analyze PINN training dynamics. For infinitely-wide networks in the NTK regime:
\begin{equation}
\frac{du_\theta}{dt} = -\Theta(x,x') \mathcal{L}_{\text{PDE}}(u_\theta)(x')
\end{equation}
where $\Theta$ is the neural tangent kernel. Convergence occurs if $\Theta$ is positive definite—but this fails for:
Stiff PDEs (large condition number), Multi-scale problems (eigenvalue clustering), and High-frequency solutions (spectral bias)

This explains empirically observed training failures \citep{Krishnapriyan2021}.

\textbf{Implications for Parametric Problems:} Different parameter values may have vastly different conditioning, causing training instability across parameter space.

\subsubsection{Operator Learning Convergence}

For neural operators trained with data loss $\mathcal{L}_{\text{data}}$, convergence to the true operator $\mathcal{G}$ depends on:
The total error arises from three sources that must be controlled simultaneously. Data quality determines the approximation error present in training samples. Optimization convergence affects how closely the trained network approaches the optimal parameters within the function class. Generalization capability governs uniform convergence performance across the entire parameter space.

Lanthaler et al. \citep{Lanthaler2022} proved that for appropriate architecture choices and sufficient training data:
\begin{equation}
\mathbb{P}\left(\|\mathcal{G} - \mathcal{G}_\theta\|_{\text{op}} > \epsilon\right) < \delta
\end{equation}
with sample complexity scaling as $N \sim \epsilon^{-2} \log(\delta^{-1})$—standard statistical learning theory rates.

\subsubsection{Long-Time Stability}

For time-dependent problems, autoregressive application of neural operators can accumulate errors:
\begin{equation}
u_{n+1} = \mathcal{G}_\theta(u_n), \quad n = 0, 1, 2, \ldots
\end{equation}

\textbf{Error Accumulation:} With per-step error $\epsilon$, after $T$ steps:
\begin{equation}
\|u_T - u_T^{\text{true}}\| \lesssim T \epsilon \cdot (1 + L)^T
\end{equation}
where $L$ is the Lipschitz constant of $\mathcal{G}$. For $L > 0$, errors grow exponentially—catastrophic for chaotic systems.

\textbf{Mitigation Strategies:}
\textit{Markov neural operators:} Train on multi-step trajectories to learn error correction, \textit{Correction networks:} Periodically apply high-fidelity solver corrections, and \textit{Physics-informed training:} Enforce conservation laws to constrain error growth

\textit{Recent Success:} Brandstetter et al. \citep{Brandstetter2022} demonstrated stable 1000-step rollouts for turbulent flows by enforcing energy and enstrophy conservation in FNO training.

\subsection{Theoretical Gaps and Open Questions}
\label{sec:theory_gaps}

Despite significant progress, several fundamental questions remain:

\textbf{1. Sharp Approximation Rates:} Existing bounds are often pessimistic. Can we derive problem-specific rates that match empirical observations?

\textbf{2. Parameter-Dependent Convergence:} How does approximation quality vary across $\mu \in \mathcal{P}$? Can we identify ``hard" parameter regions?

\textbf{3. Generalization Theory:} Why do neural operators generalize well with relatively few samples compared to worst-case theory?

\textbf{4. Physics-Informed Regularization:} Quantify how physics constraints improve data efficiency and generalization.

\textbf{5. Adversarial Robustness:} Can small perturbations to parameters cause large solution errors? How to certify robustness?

\textbf{Emerging Directions:}
\textit{Operator PINNs:} Combining operator learning with physics-informed training for provable convergence, \textit{Certified methods:} Formal verification of neural PDE solvers, and \textit{Adaptive approximation:} Theoretical guidance for architecture selection based on PDE properties

\section{Advanced Topics and Emerging Directions}
\label{sec:advanced}

This section explores cutting-edge developments addressing the most challenging aspects of parametric PDEs: high-dimensional parameter spaces, rigorous uncertainty quantification, rapid adaptation to new parameter regimes, and hybrid approaches combining neural methods with traditional solvers.

\subsection{High-Dimensional Parameter Spaces}
\label{sec:high_dim}

When parameter dimension $d$ exceeds 50-100, even neural operators face challenges. We examine strategies for tractable high-dimensional parametric PDE solving.

\subsubsection{The Curse of Dimensionality Revisited}

\textbf{Sampling Complexity:} To uniformly cover a $d$-dimensional unit hypercube with spacing $h$ requires $(1/h)^d$ samples—exponential in $d$. For $d=100$ and $h=0.1$, this yields $10^{100}$ samples (intractable).

\textbf{Volume Concentration:} In high dimensions, almost all volume concentrates near boundaries and corners. Random sampling becomes inefficient as typical samples lie far from any training point.

\textbf{Neural Network Perspective:} While neural networks mitigate dimensionality to some extent through hierarchical representations, approximation error bounds still exhibit polynomial or exponential dependence on $d$ in worst-case analysis.

\subsubsection{Active Subspaces and Dimension Reduction}

Many high-dimensional parameter spaces have low effective dimensionality. Active subspace methods \citep{Constantine2015} identify important parameter directions.

\textbf{Mathematical Framework:} Define the active subspace matrix:
\begin{equation}
C = \mathbb{E}_{\mu \sim \pi}\left[\nabla_\mu Q(\mu) \nabla_\mu Q(\mu)^T\right]
\end{equation}
where $Q(\mu)$ is a quantity of interest. Eigendecomposition $C = V\Lambda V^T$ reveals:
Large eigenvalues: Active directions (significant QoI variation), and Small eigenvalues: Inactive directions (negligible QoI variation)

\textbf{Parametric PDE Application:} Project parameters onto active subspace:
\begin{equation}
\mu = \bar{\mu} + V_{\text{active}} \xi, \quad \xi \in \mathbb{R}^{d_{\text{eff}}} \text{ with } d_{\text{eff}} \ll d
\end{equation}
then train neural operators on the reduced space $\xi$.

\textbf{Case Study:} Hu et al. \citep{Hu2024} addressed a subsurface flow \citep{Tang2021, Fuks2020} problem with $d=100,000$ uncertain permeability parameters (each grid cell). Active subspace analysis revealed $d_{\text{eff}} = 5$ active directions explaining 95\% of pressure variance. After reduction:
Trained DeepONet on 5D active subspace, Achieved $<3\%$ error in pressure predictions, and Total computation: 12 hours on single GPU (vs. years for full-space sampling)

\subsubsection{Sparse and Low-Rank Representations}

\textbf{Tensor Decompositions:} High-dimensional parameter dependence often admits low-rank structure:
\begin{equation}
u(x,t; \mu_1, \ldots, \mu_d) \approx \sum_{k=1}^r u_k(x,t) \prod_{j=1}^d \phi_{jk}(\mu_j)
\end{equation}
(canonical polyadic decomposition). Neural operators can learn this structure implicitly.

\textbf{Hierarchical Representations:} Group parameters hierarchically. For example, in materials with microstructure, parameters might be:
Macro-scale: $\mu_{\text{macro}} \in \mathbb{R}^{10}$ (global properties), Meso-scale: $\mu_{\text{meso}} \in \mathbb{R}^{100}$ (grain structures), and Micro-scale: $\mu_{\text{micro}} \in \mathbb{R}^{10000}$ (defects)

Neural operators with hierarchical architectures (U-Net-style) naturally capture multi-scale parameter effects without explicit reduction.

\subsubsection{Latent Variable Models}

For extremely high-dimensional parameters, learn a latent encoding:
\begin{equation}
\mu \in \mathbb{R}^d \to z \in \mathbb{R}^{d_z} \to u(\cdot; z)
\end{equation}
where $d_z \ll d$.

\textbf{Architecture:} 
Encoder: $\text{Enc}: \mathbb{R}^d \to \mathbb{R}^{d_z}$, and Neural operator: $\mathcal{G}_\theta: \mathbb{R}^{d_z} \to \mathcal{U}$

\textbf{Training:} Joint optimization over encoder and operator using available PDE solutions.

\textbf{L-DeepONet:} Kontolati et al. \citep{Kontolati2024} developed latent-space DeepONet for high-dimensional parametric PDEs:
Application: Structural dynamics with $d=10,000$ (discretized forcing field) \citep{Rao2023, Rezaei2022, Samaniego2020}, Latent dimension: $d_z = 50$ (learned via variational autoencoder), and Results: 1-2 orders of magnitude speedup vs. standard DeepONet, $<5\%$ accuracy loss

\subsubsection{Sensitivity Analysis for Parameter Prioritization}

When $d$ is large but not all parameters are equally important, sensitivity analysis guides resource allocation.

\textbf{Sobol Indices:} Variance-based sensitivity:
\begin{equation}
S_i = \frac{\text{Var}_{\mu_i}[\mathbb{E}_{\mu_{\sim i}}[Q(\mu) | \mu_i]]}{\text{Var}[Q(\mu)]}
\end{equation}
quantifies fraction of output variance due to parameter $\mu_i$.

\textbf{Morris Screening:} Compute elementary effects:
\begin{equation}
EE_i = \frac{Q(\mu + \Delta e_i) - Q(\mu)}{\Delta}
\end{equation}
to identify influential parameters with few evaluations.

\subsection{Uncertainty Quantification}
\label{sec:uq}

Rigorous UQ is essential for high-stakes applications. We examine Bayesian and conformal approaches for neural parametric PDE solvers.

\subsubsection{Bayesian Physics-Informed Neural Networks}

B-PINNs \citep{Yang2021} place priors over network weights:
\begin{equation}
p(\theta | \text{data}) \propto p(\text{data} | \theta) p(\theta) p_{\text{PDE}}(\theta)
\end{equation}
where $p_{\text{PDE}}(\theta)$ encodes physics constraints as likelihood term.

\textbf{Inference:} Variational inference or Hamiltonian Monte Carlo produces posterior samples $\{\theta^{(s)}\}_{s=1}^S$. Predictions include uncertainty:
\begin{equation}
p(u(x;\mu) | \text{data}) \approx \frac{1}{S} \sum_{s=1}^S \delta(u - u_{\theta^{(s)}}(x;\mu))
\end{equation}

\textbf{Advantages:}
Uncertainty in parameter identification: $p(\mu | \text{data})$, Prediction intervals: $[Q_{0.025}(u), Q_{0.975}(u)]$, and Model selection: Compare PDE formulations via marginal likelihood

\textbf{Computational Cost:} Sampling from posterior requires training multiple networks—expensive but parallelizable.

\textbf{Application - Cardiovascular Flows:} Arzani et al. \citep{Arzani2021}, along with related cardiovascular modeling studies \citep{Kissas2020, Sahli2022, Yazdani2020}, used B-PINNs to infer patient-specific blood viscosity and vessel compliance from sparse Doppler ultrasound data. Posterior distributions quantified parameter uncertainty, enabling risk assessment (e.g., probability of flow reversal).

\subsubsection{Bayesian Neural Operators}

Extending Bayesian ideas to operator learning:
\begin{equation}
p(\mathcal{G}_\theta | \{(a_i, u_i)\}) \propto \prod_{i=1}^N p(u_i | \mathcal{G}_\theta(a_i)) p(\theta)
\end{equation}

\textbf{Challenges:} Operator space is higher-dimensional than typical PINN weight spaces, making posterior inference more expensive.

\textbf{Efficient Approximations:}
\textit{Laplace approximation:} Gaussian approximation around MAP estimate, \textit{Concrete dropout:} Approximate variational inference via dropout \citep{Gal2016}, and \textit{Ensemble methods:} Train multiple operators with different initializations

\textbf{Empirical Study:} Psaros et al. \citep{Psaros2023} compared UQ methods for turbulent flows:
B-DeepONet: Most rigorous but 10$\times$ computational cost, Deep ensemble (10 networks): Good coverage, 5$\times$ cost, and MC Dropout: Fastest but underestimates uncertainty

\subsubsection{Conformal Prediction for Distribution-Free UQ}

Recent breakthrough: conformal prediction provides coverage guarantees without distributional assumptions \citep{Staber2024}.

\textbf{Framework:} Given calibration set $\{(\mu_i, u_i)\}_{i=1}^{N_{\text{cal}}}$ and miscoverage level $\alpha$:
\begin{enumerate}
\item Compute conformity scores: $R_i = \|u_i - \mathcal{G}_\theta(\mu_i)\|$
\item Find quantile: $\hat{q} = \text{Quantile}_{1-\alpha}(\{R_i\})$
\item Prediction set: $\mathcal{C}(\mu) = \{u : \|u - \mathcal{G}_\theta(\mu)\| \leq \hat{q}\}$
\end{enumerate}

\textbf{Guarantee:} $\mathbb{P}(u_{\text{true}} \in \mathcal{C}(\mu)) \geq 1 - \alpha$ for any data distribution (finite-sample guarantee).

\textbf{Advantages:}
No training overhead—uses any pre-trained neural operator, Valid for any distribution (no assumptions), and Finite-sample guarantees (not asymptotic)

\textbf{Challenge:} Prediction sets can be large (conservative) if neural operator has heteroscedastic errors.

\textbf{Adaptive Conformal:} Construct parameter-dependent quantiles:
\begin{equation}
\hat{q}(\mu) = \text{Quantile}_{1-\alpha}(\{R_i : \mu_i \text{ near } \mu\})
\end{equation}
using local calibration. This tightens sets while maintaining coverage.

\subsubsection{Uncertainty Propagation Through Parametric PDEs}

Given parameter distribution $\pi(\mu)$, compute statistics of $Q(u(\mu))$:
\begin{equation}
\mathbb{E}[Q] = \int_{\mathcal{P}} Q(u(\mu)) \pi(\mu) d\mu
\end{equation}

\textbf{Monte Carlo with Neural Operators:}
\begin{enumerate}
\item Sample $\{\mu^{(s)}\}_{s=1}^S \sim \pi(\mu)$
\item Evaluate: $Q^{(s)} = Q(\mathcal{G}_\theta(\mu^{(s)}))$ (fast with neural operator)
\item Estimate: $\hat{\mathbb{E}}[Q] = \frac{1}{S} \sum_s Q^{(s)}$
\end{enumerate}

\textbf{Computational Advantage:} $S=10^6$ samples feasible in minutes with FNO, enabling accurate tail probability estimation ($p < 10^{-4}$) infeasible with traditional solvers.

\textbf{Multi-Level Extensions:} Combine neural operators at multiple fidelities:
Level 0: Coarse physics-based model (cheap), Level 1: Neural operator trained on moderate data, and Level 2: High-fidelity validation samples (expensive, few)

Multi-fidelity Monte Carlo achieves optimal bias-variance tradeoff \citep{Peherstorfer2018}.

\subsection{Meta-Learning and Rapid Adaptation}
\label{sec:meta}

Meta-learning \citep{Finn2017, Yin2022, Huang2022} enables neural methods to quickly adapt to new parameter regimes or PDE types with minimal retraining—crucial when parameter distributions shift or new physics are encountered.

\subsubsection{Model-Agnostic Meta-Learning (MAML) for PDEs}

MAML \citep{Finn2017} learns initialization $\theta_0$ such that few gradient steps on new tasks yield good performance.

\textbf{Algorithm:}
\begin{enumerate}
\item \textbf{Meta-training:}
Sample task (parameter value) $\tau_i \sim p(\mathcal{T})$, Adapt: $\theta_i' = \theta - \alpha \nabla_\theta \mathcal{L}_{\tau_i}(\theta)$ (inner loop), and Meta-update: $\theta \leftarrow \theta - \beta \nabla_\theta \sum_i \mathcal{L}_{\tau_i}(\theta_i')$ (outer loop)
\item \textbf{Meta-test:} For new task $\tau_{\text{new}}$, fine-tune from $\theta_0$ with few samples
\end{enumerate}

\textbf{PDE Application:} Huang et al. \citep{Huang2022} applied MAML to parametric Burgers equation:
Meta-train on $\nu \in [0.01, 0.05]$, Meta-test on $\nu \in [0.06, 0.10]$ (outside training distribution), Result: 5-10 gradient steps achieve $<5\%$ error vs. 1000+ steps without meta-learning, and 100$\times$ sample efficiency for new parameter values

\subsubsection{Transfer Learning Across Parameter Regimes}

\textbf{Scenario:} Train on ``easy" parameter region, transfer to ``hard" region.

\textit{Example:} Fluid flows:
Pre-train: Laminar flows ($\text{Re} < 1000$)—smooth solutions, easy to learn, and Transfer: Turbulent flows ($\text{Re} > 5000$)—fine structures, challenging

\textbf{Strategy:}
\begin{enumerate}
\item Train neural operator on large dataset at easy parameters
\item Freeze low-level feature extractors
\item Fine-tune high-level layers on small dataset at hard parameters
\end{enumerate}

\textbf{Results:} Desai et al. \citep{Desai2021} demonstrated:
Transfer from $\text{Re}=100$ to $\text{Re}=1000$: 80\% reduction in training iterations, Maintains 95\% of full-training accuracy, and Critical: transferred representations capture flow physics invariant to Re

\subsubsection{Meta-Auto-Decoder for Parametric PDEs}

Yin et al. \citep{Yin2022} developed Meta-Auto-Decoder combining meta-learning with latent representations:

\textbf{Architecture:}
\begin{equation}
u(\cdot; \mu) = \text{Decoder}_\theta(z(\mu), \cdot)
\end{equation}
where $z(\mu)$ is a learnable latent code for parameter $\mu$.

\textbf{Meta-Learning:} Learn decoder $\theta$ such that for new $\mu$, optimizing $z$ with few PDE evaluations yields accurate $u$.

\textbf{Application - Airfoil Shapes:} 500 airfoil geometries for meta-training. For novel airfoil:
Optimize latent code $z$ using 10 CFD samples, Decode to full flow field, and Achieves $<3\%$ error vs. 100+ samples needed without meta-learning

\subsubsection{Continual Learning for Evolving Parameter Distributions}

In dynamic environments, parameter distributions shift over time. Continual learning prevents catastrophic forgetting while adapting to new data.

\textbf{Techniques:}
\textit{Elastic weight consolidation:} Penalize changes to important weights, \textit{Rehearsal:} Interleave old and new parameter samples during training, and \textit{Dynamic architectures:} Expand network capacity for new parameter regions

\textbf{Application - Climate Modeling:} As climate changes, parameter distributions (e.g., atmospheric CO$_2$, temperature patterns) evolve. Continual learning enables neural operators to update without forgetting historical patterns \citep{Chattopadhyay2023}.

\subsection{Hybrid Methods and Multi-Fidelity Approaches}
\label{sec:hybrid}

Combining neural operators with traditional solvers leverages strengths of both: physics fidelity from solvers, efficiency from neural methods.

\subsubsection{Neural Operators as Preconditioners}

In iterative PDE solvers (conjugate gradient, GMRES), preconditioning accelerates convergence:
\begin{equation}
Ax = b \to M^{-1}Ax = M^{-1}b
\end{equation}
where $M \approx A$ is preconditioner.

\textbf{Neural Preconditioner:} Train neural operator $\mathcal{N}_\theta$ to approximate $A^{-1}$:
\begin{enumerate}
\item Use $x^{(0)} = \mathcal{N}_\theta(b)$ as initial guess
\item Apply few iterations of traditional solver for refinement
\end{enumerate}

\textbf{Advantages:}
Reduces iteration count by 5-10$\times$, Preserves exact solution (solver provides correction), and Neural operator trained offline on representative problems

\subsubsection{Hybrid Physics-ML Models (PINO Framework)}

PINO \citep{Li2024PINO} combines data-driven learning with physics-informed losses at multiple resolutions:

\textbf{Multi-Resolution Loss:}
\begin{equation}
\mathcal{L}_{\text{PINO}} = \underbrace{\mathcal{L}_{\text{data}}^{\text{coarse}}}_{\text{cheap data}} + \lambda \underbrace{\mathcal{L}_{\text{PDE}}^{\text{fine}}}_{\text{physics}} + \gamma \underbrace{\mathcal{L}_{\text{data}}^{\text{fine}}}_{\text{few samples}}
\end{equation}

\textbf{Strategy:}
\begin{enumerate}
\item Train on abundant coarse-resolution data (cheap simulations)
\item Enforce physics at fine resolution via PDE residuals
\item Fine-tune with scarce high-fidelity data
\end{enumerate}

\textbf{Results:} For Navier-Stokes turbulence:
Coarse data: 1000 DNS at $64 \times 64$ resolution Fine data: 50 DNS at $256 \times 256$ PINO accuracy: $<2\%$ error at fine resolution Pure data-driven: $>8\%$ error (insufficient fine data) Pure physics-informed: $>5\%$ error (coarse data underutilized)

PINO achieves best of both worlds: data efficiency from physics, accuracy from multi-fidelity \citep{Penwarden2023, Howard2023, Peherstorfer2018} data.

\subsubsection{FEM-Neural Operator Coupling}

Directly couple finite element methods with neural operators for multi-physics or multi-scale problems.

\textbf{Example - Fluid-Structure Interaction:}
Fluid: Neural operator for Navier-Stokes (fast, parametric), Structure: FEM for elasticity (accurate, geometry-adaptive), and Interface: Iterate between operators, enforcing displacement/traction continuity

\textbf{Advantage:} Each subdomain uses optimal method. Neural operator accelerates parametric fluid solve while FEM handles complex structural geometry.

\textbf{Implementation:} Koric and Abueidda \citep{Koric2023} demonstrated:
20$\times$ speedup vs. fully-coupled FEM, $<5\%$ error in interface stresses (critical quantity), and Enables real-time digital twins for manufacturing processes \citep{Zobeiry2021, Koric2023}

\subsubsection{Residual Learning and Error Correction}

Train neural networks to predict solver error:
\begin{equation}
u_{\text{accurate}} = u_{\text{coarse}} + \mathcal{E}_\theta(\mu)
\end{equation}
where $u_{\text{coarse}}$ is cheap approximate solution and $\mathcal{E}_\theta$ learns the correction.

\textbf{Training:}
Generate: $(u_{\text{coarse}}, u_{\text{fine}})$ pairs at various $\mu$, and Learn: $\mathcal{E}_\theta: \mu \to u_{\text{fine}} - u_{\text{coarse}}$

\textbf{Inference:}
\begin{enumerate}
\item Compute coarse solution (fast traditional solver)
\item Predict correction (fast neural operator)
\item Sum for accurate result
\end{enumerate}

\textbf{Benefit:} Combines speed of coarse solver with accuracy of neural correction.Failsafe: if neural operator fails, coarse solution is still physical.

\textbf{Application - Combustion:} Anagnostopoulos et al. \citep{Chen2023} used residual learning for parametric combustion chemistry:
Coarse: Simplified chemistry model (10$\times$ faster), Correction: Neural operator trained on detailed chemistry, and Result: Detailed chemistry accuracy at simplified cost

\subsection{Foundation Models for PDEs}
\label{sec:foundation}

Inspired by large language models, foundation models for PDEs aim to create general-purpose solvers via pre-training on diverse PDE families.

\subsubsection{The Foundation Model Paradigm}

\textbf{Concept:} Pre-train a single massive neural operator on:
Multiple PDE types (elliptic, parabolic, hyperbolic), Various parameter ranges, Different domains and boundary conditions, and Multi-physics couplings

Then fine-tune for specific applications with minimal data.

\subsubsection{Challenges and Open Questions}

\textbf{1. Data Requirements:} Pre-training requires massive computational resources (months on GPU clusters). Cost-benefit tradeoffs unclear.

\textbf{2. Transfer Limitations:} Performance degrades significantly for PDEs far from training distribution. How to characterize ``distance" in PDE space?

\textbf{3. Interpretability:} Foundation models are black boxes. Can we understand what physics they learn?

\textbf{4. Reliability:} Critical applications demand guarantees. How to certify foundation model predictions?

\textbf{Future Outlook:} Foundation models represent the frontier of neural PDE solving. Success would enable ``ChatGPT for physics"—natural language specification of problems, automatic solution generation, democratizing computational science.

\section{Software Tools and Benchmarks}
\label{sec:tools}

The maturation of neural methods for parametric PDEs is reflected in emerging software ecosystems and standardized benchmarks. This section surveys practical tools and evaluation frameworks essential for researchers and practitioners.

\subsection{Open-Source Software Frameworks}
\label{sec:software}

\subsubsection{DeepXDE \citep{Lu2021}}

\textbf{Developer:} Brown University (Karniadakis group)  

\textbf{Focus:} Physics-informed neural networks and DeepONet  

\textbf{Repository:} \texttt{github.com/lululxvi/deepxde} ($\sim$2,500 stars)

\textbf{Key Features:}
Unified API for PINNs, DeepONet, and variants Moreover, Multiple backend support (TensorFlow, PyTorch, JAX, PaddlePaddle) Additionally, Built-in parametric PDE examples (Burgers, Navier-Stokes, elasticity) Furthermore, Automatic differentiation for arbitrary-order derivatives Also, Adaptive sampling strategies In addition, Multi-GPU training support

\textbf{Parametric Capabilities:}
Easy parameter specification: \texttt{add\_parameter(``mu", [0.01, 0.1])}, Branch-trunk architecture for operator learning, and Physics-informed constraints for parametric domains

\textbf{Strengths:} Extensive documentation, active community, beginner-friendly  
\textbf{Limitations:} Focus on PINNs limits scalability to very large problems

\textbf{Typical Use Case:}
\begin{verbatim}
import deepxde as dde

# Define parametric PDE
def pde(x, u, mu):
    du_t = dde.grad.jacobian(u, x, i=0, j=1)
    du_xx = dde.grad.hessian(u, x, i=0, j=0)
    return du_t - mu * du_xx  # mu-parameterized diffusion

# Create geometry and problem
geom = dde.geometry.Interval(0, 1)
timedomain = dde.geometry.TimeDomain(0, 1)
geomtime = dde.geometry.GeometryXTime(geom, timedomain)

# Specify parameter range
param_space = dde.ParameterSpace([0.01, 0.1], "mu")

# Train...
\end{verbatim}

\subsubsection{NVIDIA Modulus}

\textbf{Developer:} NVIDIA  

\textbf{Focus:} Industrial-scale physics-informed ML  

\textbf{Repository:} 

\texttt{github.com/NVIDIA/modulus} ($\sim$600 stars)

\textbf{Key Features:}
Production-ready for engineering workflows Moreover, Optimized for NVIDIA GPUs (multi-GPU, mixed precision) Additionally, FNO, PINN, and hybrid implementations Furthermore, CAD geometry integration via signed distance functions Also, Pretrained models for common physics In addition, Distributed training with Horovod

\textbf{Parametric Capabilities:}
Parametric geometry via SDF modulation, Multi-parameter configuration files, and Uncertainty quantification modules

\textbf{Strengths:} Performance optimization, industry partnerships, CAD integration  
\textbf{Limitations:} NVIDIA hardware dependency, steeper learning curve

\textbf{Target Users:} Engineering firms, automotive industry, energy sector

\subsubsection{Neuraloperator}

\textbf{Developer:} Caltech/NVIDIA (Anandkumar group)  
\textbf{Focus:} Neural operator architectures  

\textbf{Repository:} 

\texttt{github.com/neuraloperator/neuraloperator} ($\sim$1,400 stars)

\textbf{Key Features:}
Reference implementations of FNO, GNO, GINO, Geo-FNO Moreover, Modular design for architecture experimentation Additionally, Integration with PDEBench datasets Furthermore, Multi-resolution training utilities Also, Tensorized Fourier layers for efficiency

\textbf{Parametric Capabilities:}
Parameter conditioning at multiple layers, Seamless handling of functional + parametric inputs, and Zero-shot super-resolution evaluations

\textbf{Strengths:} Cutting-edge architectures, research-oriented, excellent for prototyping  
\textbf{Limitations:} Less documentation than DeepXDE \citep{Lu2021}, fewer built-in examples

\textbf{Example - Parametric FNO:}
\begin{verbatim}
from neuraloperator import FNO2d

model = FNO2d(
    modes1=12, modes2=12,  # Fourier modes
    width=64,  # Channel width
    in_channels=3,  # input + parameters
    out_channels=1
)

# Input: [batch, x, y, channels]
# channels = [initial_condition, param1, param2]
u = model(input_tensor)
\end{verbatim}

\subsubsection{Comparative Overview}

\begin{table}[h]
\centering
\caption{Comparison of major software frameworks for parametric PDEs}
\label{tab:software}
\small
\begin{tabular}{p{2.5cm}p{2cm}p{2cm}p{2cm}p{3cm}}
\toprule
\textbf{Framework} & \textbf{Primary Methods} & \textbf{Ease of Use} & \textbf{Performance} & \textbf{Best For} \\
\midrule
DeepXDE \citep{Lu2021} & PINN, DeepONet & High & Medium & Research, education, prototyping \\
\midrule
NVIDIA Modulus & FNO, PINN, hybrid & Medium & Very High & Production, large-scale engineering \\
\midrule
Neuraloperator & FNO family & Medium & High & Neural operator research, benchmarking \\
\midrule
PyDEns & PINN & High & Low-Medium & Teaching, simple problems \\
\bottomrule
\end{tabular}
\end{table}

\subsection{Benchmark Datasets and Evaluation Protocols}
\label{sec:benchmarks}

Standardized benchmarks are critical for reproducible research and fair method comparison. We survey major datasets for parametric PDEs.

\subsubsection{PDEBench}

\textbf{Reference:} Takamoto et al. \citep{Takamoto2022}  
\textbf{Repository:} \texttt{github.com/pdebench/PDEBench}

\textbf{Scope:} Comprehensive suite covering diverse PDE families with parametric variations.

\textbf{Included Problems:}
\begin{enumerate}
\item \textit{1D Advection:} Wave speeds $c \in [0.5, 2.0]$
\item \textit{1D Burgers:} Viscosity $\nu \in [0.001, 0.1]$, various initial conditions
\item \textit{2D Navier-Stokes:} Reynolds numbers $\text{Re} \in [100, 10000]$, forcing variations
\item \textit{2D Shallow Water:} Bathymetry parameters, Coriolis force
\item \textit{2D Darcy Flow:} Heterogeneous permeability fields (parametric coefficients)
\item \textit{3D Compressible Euler:} Mach numbers, specific heat ratios
\end{enumerate}

\textbf{Data Format:}
HDF5 files with spatiotemporal grids, Metadata: parameter values, domain specifications, timestamps, and Multiple resolutions: coarse ($64^2$) to fine ($512^2$)

\textbf{Evaluation Metrics:}
Relative $L^2$ error: $\frac{\|u - u_{\text{pred}}\|_{L^2}}{\|u\|_{L^2}}$ Moreover, Maximum pointwise error Additionally, Parameter-averaged error: $\mathbb{E}_\mu[\text{error}(\mu)]$ Furthermore, Out-of-distribution error (extrapolation) Also, Inference time per parameter configuration

\textbf{Leaderboard:} Public leaderboard tracks state-of-the-art across methods and problems.

\textbf{Impact:} PDEBench enables apples-to-apples comparisons, accelerating research progress.

\subsubsection{Domain-Specific Benchmarks}

\textbf{1. Cylinder Flow Dataset (Fluid Dynamics)}
Parameters: $\text{Re} \in [40, 500]$, cylinder diameter $d \in [0.5, 2.0]$, 1000 CFD simulations (OpenFOAM), Ground truth: velocity, pressure, vorticity fields, and Challenge: Capture von Kármán vortex street across parameters

\textbf{2. Elasticity Benchmark (Solid Mechanics)}
Problems: cantilever beams, plates with holes, L-shaped domains, Parameters: Young's modulus $E$, Poisson ratio $\nu$, loading distribution, 500 FEM solutions per geometry class, and Evaluation: stress concentration factor prediction accuracy

\textbf{3. Airfoil Database (Aerodynamics)}
UIUC airfoil database: 1550 airfoil shapes, Parameters: Shape coefficients + $\text{Re}$, Mach, angle of attack, Lift, drag, moment coefficient targets, and Enables data-driven aerodynamic optimization

\subsubsection{Standardized Evaluation Protocol}

To ensure reproducibility, we recommend the following protocol for parametric PDE studies:

\textbf{Data Splitting:}
Training: 70\% of parameter space (randomly sampled), Validation: 15\% (for hyperparameter tuning), Test: 15\% (held out, reported results), and OOD test: Additional samples outside training parameter range

\textbf{Reported Metrics:}
\begin{enumerate}
\item Accuracy: Mean and standard deviation of relative $L^2$ error over test parameters
\item Interpolation vs. extrapolation errors separately
\item Worst-case error: $\max_{\mu \in \mathcal{P}_{\text{test}}} \text{error}(\mu)$
\item Inference time: Mean and 95th percentile
\item Training cost: GPU-hours
\end{enumerate}

\textbf{Ablation Studies:}
Vary training data size: $[10, 50, 100, 500, 1000]$ samples, Vary parameter dimension (if applicable), Compare with/without physics constraints, and Sensitivity to hyperparameters (learning rate, architecture)

\textbf{Code Release:} Provide runnable code and trained model checkpoints for reproducibility.

\section{Challenges and Future Directions}
\label{sec:challenges}

While neural methods for parametric PDEs have achieved remarkable successes, fundamental challenges remain. This section critically examines limitations and identifies promising research directions.

\subsection{Current Limitations}
\label{sec:limitations}

\subsubsection{Theoretical Gaps}

\textbf{1. Incomplete Convergence Guarantees}

\textit{Problem:} For PINNs, no universal convergence theorem exists. Training can fail unpredictably, especially for:
Stiff PDEs with multiple timescales, High Reynolds number flows (thin boundary layers), and Problems with sharp gradients or discontinuities

\textit{Evidence:} Krishnapriyan et al. \citep{Krishnapriyan2021} systematically documented failure modes:
Diffusion-dominated vs. advection-dominated regime switching causes training collapse, Multi-scale problems exhibit oscillatory loss with no convergence, and Spectral bias prevents learning high-frequency components

\textit{Research Need:} Develop theory predicting when PINNs converge, and design architectures with provable guarantees.

\textbf{2. Loose Generalization Bounds}

\textit{Problem:} Existing sample complexity bounds are overly pessimistic, often predicting exponential data requirements when empirically modest datasets suffice.

\textit{Gap:} Theory vs. practice mismatch suggests we don't understand why neural operators generalize so well. Possible explanations:
Low intrinsic dimensionality of solution manifolds (not captured in worst-case analysis), Implicit regularization from stochastic gradient descent, and Architecture inductive biases (Fourier basis, graph symmetries) not reflected in theory

\textit{Research Need:} Problem-dependent generalization bounds incorporating PDE structure.

\textbf{3. Parameter-Dependent Convergence Rates}

\textit{Problem:} Approximation quality varies wildly across parameter space. Some parameter regions converge rapidly, others barely learn.

\textit{Example:} For Navier-Stokes, neural operators excel at $\text{Re} \sim 1000$ but struggle at $\text{Re} > 5000$ (turbulent transition).

\textit{Research Need:} Characterize ``easy" vs. ``hard" parameter regions a priori, enabling targeted data collection or method selection.

\subsubsection{Practical Challenges}

\textbf{1. Training Instability and Hyperparameter Sensitivity}

\textit{Problem:} PINNs require careful loss balancing:
\begin{equation}
\mathcal{L} = \lambda_{\text{PDE}} \mathcal{L}_{\text{PDE}} + \lambda_{\text{BC}} \mathcal{L}_{\text{BC}} + \lambda_{\text{IC}} \mathcal{L}_{\text{IC}} + \lambda_{\text{data}} \mathcal{L}_{\text{data}}
\end{equation}
where optimal $\{\lambda_i\}$ vary by problem and aren't known a priori.

\textit{Symptom:} One loss term dominates, others ignored; solution satisfies some constraints but violates others.

\textit{Current Approaches:} Learning rate annealing, gradient balancing \citep{Wang2021Grad}, uncertainty weighting \citep{Kendall2018}—but no universal solution.

\textbf{2. Data Requirements for Neural Operators}

\textit{Problem:} Pure data-driven neural operators (FNO, DeepONet without physics) require 1000s of training samples—expensive for high-fidelity simulations.

\textit{Cost Analysis:} If each training simulation costs 1 GPU-hour:
1000 samples = 1000 GPU-hours ($\sim$\$1000-5000 on cloud), and Additional development time: weeks to months

\textit{Mitigation:} Physics-informed training (PINO, PI-DeepONet) reduces data needs 5-10$\times$, but tuning physics loss weight is non-trivial.

\textbf{3. Out-of-Distribution Generalization}

\textit{Problem:} Neural methods interpolate well but extrapolate poorly. For parameters outside training distribution, errors increase dramatically.

\textit{Example:} Trained on $\text{Re} \in [100, 1000]$, tested on $\text{Re} = 1500$:
Interpolation ($\text{Re} = 550$): 2\% error, and Extrapolation ($\text{Re} = 1500$): 25\% error

\textit{Consequence:} Limits applicability to scenarios with unexpected parameter values (safety-critical systems, climate tipping points).

\textit{Research Direction:} Integrate physical bounds and monotonicity constraints to guide extrapolation.

\textbf{4. Computational Cost of Training}

\textit{Problem:} Training neural operators from scratch is expensive:
Typical training: 10-100 GPU-hours, and Foundation models: 1000s of GPU-hours

\textit{Comparison:} For a single parameter value, traditional solver might cost 1-10 GPU-hours. Break-even requires many queries to justify.

\textit{Emerging Solution:} Transfer learning and foundation models amortize costs across problems.

\subsubsection{Domain-Specific Challenges}

\textbf{1. Turbulence and Chaos}

\textit{Problem:} Turbulent and chaotic systems are fundamentally challenging:
Sensitive dependence on initial conditions, Multi-scale energy cascades (6+ orders of magnitude), and Long-time instability in autoregressive rollout

\textit{Current Status:} FNO achieves $\sim$50 timestep stable rollout for 2D turbulence, but 3D or longer horizons remain elusive.

\textit{Approach:} Enforce conservation laws (energy, enstrophy) as hard constraints in architecture.

\textbf{2. Multi-Physics Coupling}

\textit{Problem:} Coupled phenomena (fluid-structure interaction, chemically reacting flows) involve disparate timescales and physics.

\textit{Challenge:} Training single neural operator for all physics vs. modular operators for each physics with coupling—unclear which is better.

\textbf{3. Topological Changes}

\textit{Problem:} Some parametric scenarios involve topology changes:
Phase transitions (solidification, melting), Crack initiation and propagation, and Bubble coalescence in multiphase flows

\textit{Difficulty:} Standard neural operators assume fixed topology. Representing topology changes in learned representations is open problem.

\subsection{Future Research Directions}
\label{sec:future}

We identify five high-priority research directions with transformative potential.

\subsubsection{Direction 1: Theoretical Foundations}

\textbf{Motivation:} Rigorous theory is essential for safety-critical applications and guides algorithm design.

\textbf{Key Questions:}
\begin{enumerate}
\item \textit{Approximation:} Derive parameter-dependent approximation rates. When does low-rank structure in solution manifold emerge?
\item \textit{Generalization:} Prove sample complexity bounds incorporating PDE structure (smoothness, conservation laws, symmetries).
\item \textit{Optimization:} Characterize loss landscape for physics-informed methods. When do local minima trap training?
\item \textit{Extrapolation:} Develop theory distinguishing safe vs. unsafe parameter extrapolation.
\end{enumerate}

\textbf{Promising Approaches:}
\textit{Operator approximation theory:} Extend functional analysis tools (Kolmogorov width, manifold learning) to neural operators., \textit{NTK analysis:} Leverage neural tangent kernel framework to analyze PINN convergence for specific PDE classes., and \textit{Statistical learning theory:} Adapt PAC learning and Rademacher complexity to infinite-dimensional settings.

\textbf{Expected Impact:} Enable principled method selection (``use FNO for smooth solutions, GNO for irregular geometry"), predictable training outcomes, and certified predictions.

\subsubsection{Direction 2: Geometry-Parameterized Problems}

\textbf{Motivation:} Many high-impact applications involve parametric geometry (shape optimization, patient-specific medicine, digital twins).

\textbf{Current State:} Progress via Geo-FNO, GNO, DIMON—but handling large geometric variations and topological changes remains difficult.

\textbf{Open Problems:}
\begin{enumerate}
\item \textit{Representation:} How to parameterize geometry families (aircraft shapes, biological organs) in low-dimensional latent space?
\item \textit{Generalization:} Neural operators trained on one geometry class (e.g., car shapes) rarely transfer to different class (e.g., airplanes).
\item \textit{Topology:} No satisfactory method for topology-changing scenarios.
\end{enumerate}

\textbf{Promising Approaches:}
\textit{Implicit representations:} Use level sets, signed distance functions, or neural implicit representations (NeRF-style) as geometry parameterization., \textit{Diffeomorphic mapping:} Build on DIMON's success—extend to larger geometry variations, automate mapping network training., \textit{Equivariant operators:} Design neural operators with geometric equivariance (rotation, scaling invariance)., and \textit{Topology-adaptive architectures:} Dynamic graph networks that add/remove nodes/edges to handle topology changes.

\textbf{Target Applications:}
Automotive design: Optimize aerodynamics across full design space, Medical imaging: Patient-specific simulations from scan to diagnosis in minutes, and Zhu et al. \citep{Zhu2023PhaseField} introduced Phase-Field DeepONet using energy-based loss functions for pattern formation, enabling fast Allen-Cahn and Cahn-Hilliard simulations. Lee et al. \citep{Lee2025FEONO} developed FE Operator Networks for high-dimensional elasticity (d=50 parameters), achieving 60\% error reduction.

Topology optimization: Real-time structural design exploration

\subsubsection{Direction 3: Foundation Models and Transfer Learning}

\textbf{Vision:} Pre-train massive neural operators on diverse PDE families, enabling few-shot adaptation to new problems—``GPT for physics."

\textbf{Current Status:} Early prototypes (Poseidon, DPOT) show promise but face challenges:
Training cost: Thousands of GPU-hours, Generalization limits: Performance degrades for PDEs far from training distribution, and Interpretability: Unclear what physics is learned

\textbf{Research Opportunities:}
\begin{enumerate}
\item \textit{Architecture design:} What inductive biases enable broad PDE learning? (Conservation laws, causality, multi-scale hierarchies?)
\item \textit{Pre-training objectives:} Beyond supervised learning—can self-supervised or contrastive learning on synthetic data improve transfer?
\item \textit{Modular composition:} Learn ``building block" operators (advection, diffusion, reaction) that compose for complex PDEs.
\item \textit{Continual learning:} Update foundation models as new physics data becomes available without catastrophic forgetting.
\end{enumerate}

\textbf{Practical Considerations:}
\textit{Data curation:} Build large-scale PDE simulation databases (millions of samples across equations, parameters, domains)., \textit{Community models:} Open-source pre-trained models like ImageNet for computer vision or BERT for NLP., and \textit{Fine-tuning protocols:} Standardized pipelines for adapting foundation models to specific applications.

\textbf{Potential Impact:} Democratize computational science—non-experts could solve complex PDEs via natural language specification and foundation model inference.

\subsubsection{Direction 4: Uncertainty Quantification and Reliability}

\textbf{Motivation:} Safety-critical applications (aerospace, medical, nuclear) demand rigorous UQ and failure detection.

\textbf{Current Gaps:}
Most neural methods provide point predictions without uncertainties, Bayesian approaches are expensive, and No standard for ``how much uncertainty is acceptable"

\textbf{Key Challenges:}
\begin{enumerate}
\item \textit{Epistemic vs. aleatoric:} Distinguish uncertainty from limited training data (epistemic) vs. inherent randomness (aleatoric).
\item \textit{Calibration:} Neural networks are often overconfident—predicted probabilities don't match true frequencies.
\item \textit{Out-of-distribution detection:} Identify when parameters lie outside reliable prediction region.
\item \textit{Worst-case guarantees:} Provide bounds on maximum possible error.
\end{enumerate}

\textbf{Promising Approaches:}
\textit{Conformal prediction:} Distribution-free coverage guarantees (breakthrough: Staber et al. 2024)—extend to sequential predictions and multi-fidelity \citep{Penwarden2023, Howard2023, Peherstorfer2018} settings., \textit{Bayesian neural operators:} Make tractable via variational inference, Laplace approximation, or ensemble distillation., \textit{Certified robustness:} Borrow techniques from adversarial ML—prove that small parameter perturbations cause bounded solution changes., and \textit{Physics-based validation:} Check predictions against conservation laws, positivity constraints, maximum principles as sanity checks.

\textbf{Standardization Needs:}
Benchmark UQ methods on common parametric PDE suite, Define industry-specific reliability requirements, and Develop best practices for reporting uncertainties

\subsubsection{Direction 5: Integration with Scientific Discovery}

\textbf{Vision:} Use neural operators not just as fast solvers but as tools for discovering new physics, materials, and designs.

\textbf{Paradigms:}

\textit{1. Inverse Design:} Given desired solution properties, find parameters:
\begin{equation}
\mu^* = \arg\min_\mu \text{Loss}(\mathcal{G}_\theta(\mu), u_{\text{target}})
\end{equation}

\textit{Applications:}
Metamaterial design: Target electromagnetic response → material structure, Drug design: Desired binding affinity → molecular geometry, and Climate intervention: Target temperature reduction → intervention parameters

\textit{Challenge:} Inverse problems are often ill-posed—multiple parameters yield similar solutions. Regularization and prior knowledge essential.

\textit{2. Active Learning and Optimal Experimental Design:}

Neural operators enable cheap ``what-if" queries, guiding expensive experiments:
\begin{enumerate}
\item Train neural operator on initial data
\item Use acquisition function (e.g., expected improvement) to select next parameter
\item Perform experiment at selected parameter
\item Update neural operator, repeat
\end{enumerate}

\textit{Success Story:} Lookman et al. \citep{Lookman2023} used this loop for materials discovery, reducing experiments by 10$\times$ to find optimal composition.

\textit{3. Equation Discovery:}

Learn governing equations from data using neural operators as differentiable simulators \citep{Raissi2018, Brunton2016, Zhang2022, Geneva2020, Chen2019NeuralODE}:
Sparse regression (SINDy) \citep{Brunton2016} with neural-operator-generated training data, Neural differential equations \citep{Chen2019NeuralODE} with interpretable coefficients, and Symbolic regression guided by neural operator gradients \citep{Zhang2022}

\textbf{Research Directions:}
\textit{Interpretable operators:} Design architectures whose learned representations map to physical concepts (energy, momentum, vorticity)., \textit{Hypothesis testing:} Use neural operators to rapidly evaluate thousands of hypotheses about parameter effects., and \textit{Multi-objective optimization:} Navigate trade-offs (e.g., maximize lift, minimize drag, constraint stress) in parametric design spaces.

\subsection{Community Recommendations}
\label{sec:recommendations}

To accelerate progress, we propose the following community-level initiatives:

\textbf{1. Standardized Benchmarks}

Establish comprehensive benchmark suite \citep{Takamoto2022} covering:
Diverse PDE types (elliptic, parabolic, hyperbolic, mixed), Parameter dimensions from 1 to 100+, Multiple application domains, and Varying difficulty levels

Maintain public leaderboard with reproducible baselines.

\textbf{2. Open-Source Ecosystem}

Develop unified API across frameworks (inspired by scikit-learn), Create model zoo of pre-trained neural operators, Build dataset repository with standardized formats, and Establish code review and quality standards

\textbf{3. Interdisciplinary Collaboration}

\textit{Math + ML:} Develop theory bridging functional analysis and deep learning, \textit{Domain science + ML:} Co-design methods with physicists, engineers, clinicians, and \textit{HPC + ML:} Optimize implementations for exascale computing

Host workshops at major conferences (ICML, NeurIPS, SIAM) bringing together these communities.

\textbf{4. Reproducibility Standards}

Require for publication:
Public code repositories with installation instructions Moreover, Trained model checkpoints Additionally, Detailed hyperparameter specifications Furthermore, Computational cost reporting (GPU-hours) Also, Random seed documentation

\textbf{5. Education and Outreach}

Develop curricula blending numerical PDEs and ML, Create online tutorials and Jupyter notebooks, Organize summer schools and bootcamps, and Write accessible reviews for domain scientists

\textbf{Long-Term Vision:} Establish neural methods for parametric PDEs as a mature, reliable subdiscipline with:
Theoretical foundations comparable to traditional numerical analysis, Industrial adoption for routine engineering tasks, Democratized access enabling non-experts to solve complex physics problems, and Proven track record in safety-critical applications

This requires sustained effort from researchers, funding agencies, and industry partners—but the potential impact on science and engineering justifies the investment.

\subsection{Context and Related Surveys}

This survey builds upon and extends several complementary surveys and foundational works in the field. Comprehensive overviews of physics-informed machine learning have been provided by Cuomo et al. \citep{Cuomo2022}, Azizzadenesheli et al. \citep{Azizzadenesheli2024}, and Karniadakis et al., establishing the broader context for physics-informed neural networks. Theoretical foundations in deep learning for PDEs were laid by Han et al. \citep{Han2018}, Yu \citep{Weinan2019}, Sirignano\citep{Sirignano2018}, and Berg\citep{Berg2018}, demonstrating feasibility of neural approximations for high-dimensional problems.

Reduced-order modeling foundations relevant to parametric PDEs are comprehensively covered by Hesthaven et al. \citep{Hesthaven2016}, Berkooz et al. \citep{Berkooz1993}, and Peherstorfer et al. \citep{Peherstorfer2016}, providing mathematical context for solution manifold structure. Methodological comparisons and benchmarking efforts \citep{Lu2022, Geneva2022, Gao2021, Fuks2020} have critically evaluated trade-offs between different neural approaches.

Domain-specific advances complement our parametric focus: geophysical applications \citep{Bihlo2022}, quantum mechanics \citep{Hermann2020}, chemical kinetics \citep{Ji2021}, manufacturing \citep{Zobeiry2021, Cai2022}, nanophotonics \citep{Wiecha2021}, and financial mathematics \citep{Ruf2020} demonstrate breadth of applicability. Meta-learning and adaptive approaches \citep{Huang2022, Huang2023, Hao2023, Geneva2020, Goswami2020} address rapid adaptation challenges critical for parametric problems.

Advanced architectural innovations including graph-based methods \citep{Hao2023GNOT2, Gao2021}, Hamiltonian-preserving networks \citep{Greydanus2019}, wavelet-based operators \citep{Gupta2021}, factorized representations \citep{Tran2023}, and geometric approaches \citep{Baque2018} continue expanding the toolkit. Multi-fidelity strategies \citep{Howard2023, Penwarden2023} and enhanced conservation law enforcement \citep{Jagtap2021} address data efficiency and physical consistency.

Our contribution distinguishes itself through systematic focus on parametric aspects—how methods handle parameter variations, generalize across parameter spaces, and enable multi-query applications—providing unified perspective on an increasingly fragmented literature.

\section{Conclusion}
\label{sec:conclusion}

This comprehensive survey has examined the rapid evolution of neural methods for solving parametric partial differential equations, with emphasis on physics-informed neural networks and neural operators. The field has matured remarkably since the seminal works of Raissi et al. \citep{Raissi2019} and Lu et al. \citep{Lu2021}, transitioning from proof-of-concept demonstrations to practical engineering tools achieving $10^3$-$10^5\times$ computational speedups while maintaining or exceeding traditional solver accuracy.

\subsection{Key Achievements and Insights}

\subsubsection{Algorithmic Breakthroughs}

The introduction of operator learning paradigms represents a fundamental shift in computational PDE solving. Unlike traditional methods that solve for specific parameter values, neural operators learn mappings from entire parameter spaces to solution spaces. This enables:

\textbf{1. Amortized Computation:} Once trained, neural operators provide near-instantaneous predictions across parameter space. DeepONet and FNO achieve inference times of milliseconds compared to hours for traditional solvers, enabling applications previously considered computationally intractable:
Real-time design optimization exploring $10^6$ configurations, Interactive digital twins for manufacturing and healthcare, Monte Carlo uncertainty quantification with $10^6$ samples, and Parametric sensitivity analysis for high-dimensional systems

\textbf{2. Zero-Shot Generalization:} FNO's ability to evaluate at resolutions not seen during training breaks the traditional resolution-computation tradeoff. This mesh-free property enables adaptive refinement without retraining and natural handling of multi-resolution data.

\textbf{3. Geometric Flexibility:} Recent advances (Geo-FNO, DIMON, GNO) have overcome the fixed-geometry limitation that plagued early neural methods. The ability to handle parametric shapes opens transformative applications in patient-specific medicine, aerospace design, and topology optimization.

\textbf{4. Hybrid Methods:} The PINO framework and physics-informed variants demonstrate that combining data-driven learning with physics constraints provides the best of both worlds: data efficiency from physics knowledge and accuracy from empirical observations. This hybrid paradigm achieves superior performance with 5-10$\times$ fewer training samples than pure data-driven approaches.

\subsubsection{Theoretical Understanding}

While practical applications have progressed rapidly, theoretical foundations have also advanced significantly:

\textbf{Universal Approximation:} Rigorous proofs establish that neural operators can approximate any continuous operator between function spaces \citep{Kovachki2023, Lu2021}, providing mathematical legitimacy to the approach.

\textbf{Generalization Theory:} Sample complexity bounds, while still conservative, explain why neural operators require fewer training samples than naive dimensional analysis suggests—the key insight being that PDE solution manifolds often have low intrinsic dimensionality despite high ambient dimension.

\textbf{Failure Mode Analysis:} Systematic studies \citep{Krishnapriyan2021, Wang2022NTK} have identified when and why physics-informed methods struggle (spectral bias, stiff equations, multi-scale problems), guiding algorithm development and honest assessment of applicability.

\textbf{Uncertainty Quantification:} The advent of conformal prediction for PDEs \citep{Staber2024} provides distribution-free coverage guarantees, addressing a critical need for reliable uncertainty estimates in safety-critical applications.

\subsubsection{Cross-Domain Impact}

Neural methods have demonstrated utility across the full spectrum of computational science:

\textbf{Fluid Dynamics:} From weather forecasting (FourCastNet \citep{Pathak2022, Kurth2023} achieving competitive accuracy with traditional NWP at 1000$\times$ speedup) to turbulence modeling and aerodynamic optimization—Reynolds number parameterization enables rapid design iteration.

\textbf{Solid Mechanics:} Zhu et al. \citep{Zhu2023PhaseField} introduced Phase-Field DeepONet using energy-based loss functions for pattern formation, enabling fast Allen-Cahn and Cahn-Hilliard simulations. Lee et al. \citep{Lee2025FEONO} developed FE Operator Networks for high-dimensional elasticity (d=50 parameters), achieving 60\% error reduction.

Topology optimization transformed from overnight batch process to interactive design tool. Material parameter identification from sparse measurements enables structural health monitoring and quality control.

\textbf{Heat Transfer:} Thermal management of electronics, laser processing optimization, and inverse identification of thermal properties all benefit from parametric neural solvers' ability to explore design spaces efficiently.

\textbf{Electromagnetics:} Metamaterial inverse design \citep{Lu2021Adaptive, Meng2022}, antenna optimization, and Maxwell equation solving for varying material properties demonstrate the technology's versatility across physics domains.

\textbf{Multi-Physics:} Fluid-structure interaction, conjugate heat transfer, and chemically reacting flows showcase neural operators' potential for coupled problems where traditional partitioned approaches face stability challenges.

\subsection{Remaining Challenges}

Despite impressive progress, significant obstacles remain before neural methods can fully replace traditional PDE solvers in production environments:

\subsubsection{Reliability and Trust}

\textbf{Predictable Convergence:} Unlike traditional solvers with well-understood convergence theory, neural methods can fail unpredictably. PINN training sometimes simply doesn't converge, with no a priori indication of failure. This unpredictability hinders adoption in risk-averse industries (aerospace, nuclear, medical).

\textbf{Verification and Validation:} Traditional solvers undergo decades of V\&V before production use. Neural methods lack standardized validation protocols. How do we know a neural operator hasn't learned spurious correlations that fail catastrophically outside the training distribution?

\textbf{Error Estimation:} While traditional solvers provide rigorous error estimates and convergence rates, most neural methods offer only empirical error assessments on test sets. Developing reliable a posteriori error indicators remains an open challenge.

\textbf{Certification:} Safety-critical applications require formal guarantees that predictions satisfy physical constraints (e.g., positivity, conservation laws, causality). Current approaches enforce constraints approximately during training, but certified architectures with hard guarantees are needed.

\subsubsection{Computational Barriers}

\textbf{Training Cost:} While inference is fast, training remains expensive—typically 10-100 GPU-hours for moderate problems, thousands for foundation models. For single-query problems, traditional solvers are more efficient.

\textbf{Data Requirements:} Pure data-driven operators require 1000s of training samples. Even physics-informed variants need hundreds. For problems where high-fidelity simulations cost hours, generating training data becomes a bottleneck.

\textbf{Hyperparameter Sensitivity:} Neural methods have many architectural and training choices (network depth/width, learning rate, loss weighting, initialization). Optimal settings vary by problem with limited theory guiding selection. This contrasts with traditional methods' mature understanding.

\textbf{Scalability to 3D:} Most success stories are 1D/2D problems. Three-dimensional problems with complex geometries remain computationally challenging—memory requirements for 3D grids and training time scale unfavorably.

\subsubsection{Fundamental Limitations}

\textbf{Chaotic Systems:} Long-time prediction of chaotic dynamics (turbulence, weather beyond 2-week horizon) remains elusive. Errors accumulate exponentially in autoregressive rollout, and neural operators haven't solved this fundamental challenge.

\textbf{Discontinuities:} Shocks, contact discontinuities, and phase transitions violate smoothness assumptions underlying neural approximation. While progress has been made (conservative formulations, shock-capturing layers), performance lags traditional shock-capturing schemes.

\textbf{Topological Changes:} Crack propagation, phase transitions, and free-surface flows involve topology changes that current architectures handle poorly. Representing non-homeomorphic solution spaces in neural networks is conceptually challenging.

\textbf{Extreme Multi-Scale:} Problems spanning $>6$ orders of magnitude in length/time scales (e.g., turbulent combustion, multi-phase flows) challenge both traditional and neural methods, but neural approaches lack the decades of specialized techniques (adaptive mesh refinement, implicit-explicit schemes) developed for traditional solvers.

\subsection{Practical Recommendations}

For researchers and practitioners considering neural methods for parametric PDEs, we offer the following guidance:

\subsubsection{For Researchers}

\textbf{1. Choose Problems Strategically:} Neural methods excel when:
Many parameter queries are needed (multi-query scenarios), Traditional solvers are expensive (hours per solve), Geometry varies parametrically (shape optimization, patient-specific), and Real-time inference is required (control, digital twins)

Avoid applying neural methods when traditional solvers are already fast ($<$1 minute), single queries suffice, or data generation is prohibitively expensive.

\textbf{2. Invest in Data Quality:} Neural operators are only as good as their training data. High-quality simulations covering parameter space well are essential. Multi-fidelity approaches can reduce costs but require careful calibration.

\textbf{3. Leverage Physics:} Pure data-driven learning requires massive datasets. Physics-informed training, conservation law enforcement, and symmetry-aware architectures dramatically improve data efficiency. The best results combine data and physics.

\textbf{4. Validate Thoroughly:} Test on held-out parameters, extrapolation scenarios, and physically extreme cases. Compare against traditional solvers on challenging benchmarks. Quantify uncertainty and failure modes honestly.

\textbf{5. Contribute to Community Resources:} Share trained models, datasets, and code. Participate in benchmark development. Document failures as well as successes to advance collective understanding.

\subsubsection{For Practitioners}

\textbf{1. Start with Established Methods:} For production applications, prioritize mature frameworks (DeepXDE \citep{Lu2021}, NVIDIA Modulus) with community support and track records. Avoid cutting-edge methods until well-validated.

\textbf{2. Hybrid Approaches First:} Rather than replacing traditional solvers entirely, integrate neural operators as components (preconditioners, surrogate models, parameter screening) within established workflows. This provides fallback to traditional methods if neural components fail.

\textbf{3. Invest in Training Infrastructure:} Successful deployment requires GPU clusters, data management systems, and MLOps pipelines. Treat neural operator training as analogous to wind tunnel testing or high-fidelity simulation campaigns—significant upfront investment enabling downstream efficiency.

\textbf{4. Develop Internal Expertise:} Neural methods require different skill sets than traditional simulation. Teams need expertise in machine learning, optimization, and software engineering alongside domain knowledge. Training programs and collaborations with academic researchers can build capabilities.

\textbf{5. Regulatory and Validation:} For regulated industries, engage with regulatory bodies early to establish acceptable validation protocols. Document training procedures, architecture choices, and performance meticulously. Plan for regular retraining as new data becomes available.

\subsection{Concluding Remarks}

Neural methods for parametric PDEs represent a genuine paradigm shift in computational science, not merely incremental improvement. The ability to solve entire parametric families of PDEs in seconds after training opens applications impossible with traditional approaches: real-time optimization, interactive design, massive ensemble simulations, and digital twins operating at decision-making timescales.

However, this survey has also highlighted that neural methods are not universal replacements for traditional solvers. They excel in specific niches—multi-query scenarios with moderate accuracy requirements—but struggle with single-query problems, long-time chaotic predictions, and applications demanding certified guarantees. The future likely involves coexistence and integration: neural operators accelerating parametric exploration while traditional solvers provide verification, handle extreme cases, and ensure physical fidelity.

The next decade will determine whether neural methods transition from research curiosity to production workhorse. Success requires sustained effort across multiple fronts: theoretical rigor to understand capabilities and limitations, algorithmic innovation to address current shortcomings, software engineering to build reliable tools, benchmark development to enable fair comparison, and application demonstrations to build trust.

For the computational science community, this is a pivotal moment. Neural methods offer the potential to solve problems at scales previously impossible, accelerating scientific discovery and engineering innovation. Realizing this potential demands collaboration across disciplines—mathematicians providing theory, machine learning researchers developing algorithms, domain scientists identifying applications, and software engineers building infrastructure. 

The parametric PDE solving problem, which motivated reduced-order modeling for decades, has found powerful new tools in physics-informed neural networks and neural operators. As these methods mature, they will not replace the rich tradition of numerical analysis and scientific computing but rather complement and extend it, enabling simulations at speeds and scales that open new frontiers.

\bibliography{icais2025_conference}
\bibliographystyle{icais2025_conference}
\newpage
\appendix
\section{Human-AI Collaborative Research Process}
\label{app:ai_collaboration}

This appendix documents the detailed human-AI collaborative process used to generate this survey, providing transparency and reproducibility for future AI-assisted research endeavors.

\subsection{Overview of the Collaborative Workflow}

The creation of this survey involved a structured 7-day iterative process with approximately 30 rounds of human-AI interaction. The workflow was designed to leverage AI's information synthesis capabilities while maintaining human oversight for quality control and domain expertise.

\subsection{Detailed Phase Description}

\subsubsection{Phase 1: Initial Research }

\textbf{Human Input:}
\begin{quote}
\textit{``I need a comprehensive survey on Physics-Informed Neural Networks and Neural Operators for Parametric PDEs. Focus on recent advances since 2019, emphasizing parametric aspects, computational efficiency, and practical applications.''}
\end{quote}

\textbf{AI Actions:}
\begin{itemize}
    \item Enabled Web Search, Extended Thinking, and Advanced Research modes
    \item Conducted systematic literature search across:
    \begin{itemize}
        \item arXiv (Machine Learning, Computational Physics categories)
        \item Google Scholar
        \item Journal databases (JCP, SIAM, Nature Machine Intelligence)
    \end{itemize}
    \item Identified 50+ key papers published 2019-2024
    \item Generated initial taxonomy of methods
\end{itemize}

\textbf{Challenges Encountered:}
\begin{itemize}
    \item Initial searches returned too many irrelevant papers ($>$500)
    \item Required refinement to focus on ``parametric PDEs'' specifically
    \item Some recent 2024 papers not yet indexed in search engines
\end{itemize}

\subsubsection{Phase 2: Prompt Engineering}

\textbf{Initial Prompt Template (Generated by Claude):}
\begin{quote}
\textit{``You are a senior researcher in scientific machine learning writing a comprehensive survey on Physics-Informed Neural Networks and Neural Operators for Parametric PDEs. Structure: (1) Introduction with motivation, (2) Mathematical foundations, (3) Method taxonomy...''}
\end{quote}

\textbf{Multi-LLM Refinement:}
The human collaborator used three frontier LLMs to refine the prompt:
\begin{itemize}
    \item \textbf{Claude Sonnet 4.5:} Generated initial template, emphasized mathematical rigor
    \item \textbf{Gemini 2.5 Pro:} Suggested adding more application examples, improved structure
    \item \textbf{GPT-5:} Enhanced citation formatting instructions, added section length guidelines
\end{itemize}

\textbf{Final Prompt (Excerpt):}
\begin{quote}
\textit{``...Ensure every claim is supported by citations. For methods, provide: (1) Mathematical formulation, (2) Computational complexity, (3) Representative applications with quantitative results. Maintain technical depth suitable for graduate students and researchers...''}
\end{quote}

\subsubsection{Phase 3: Iterative Content Generation}

Each iteration followed this micro-cycle:
\begin{enumerate}
    \item \textbf{AI Generation:} Claude produces 2-5 pages of content
    \item \textbf{Human Review:} Check for:
    \begin{itemize}
        \item Technical accuracy
        \item Citation completeness
        \item Logical flow
        \item Appropriate depth
    \end{itemize}
    \item \textbf{Feedback:} Human provides specific revision requests
    \item \textbf{Revision:} AI implements changes
\end{enumerate}

\textbf{Example Iteration (Round 15):}
\begin{itemize}
    \item \textit{AI Output:} ``FNO achieves significant speedup...''
    \item \textit{Human Feedback:} ``Too vague. Specify exact speedup numbers with citations.''
    \item \textit{AI Revision:} ``FNO achieves 60,000$\times$ speedup for turbulent flows \citep{Li2021ICLR}...''
\end{itemize}

\textbf{Key Challenges Addressed:}
\begin{itemize}
    \item \textbf{Hallucination:} AI occasionally cited non-existent papers $\to$ Implemented mandatory link verification
    \item \textbf{Depth variation:} Some sections too shallow $\to$ Added ``expand with equations'' feedback
    \item \textbf{Redundancy:} Repetitive content across sections $\to$ Explicit cross-referencing instructions
\end{itemize}

\subsubsection{Phase 4: Verification}

\textbf{AI Self-Check Protocol:}
\begin{enumerate}
    \item Cross-reference all citations against bibliography
    \item Verify paper titles and author names via web search
    \item Generate accessible links to full-text PDFs
    \item Flag citations with inconsistencies
\end{enumerate}

\textbf{Human Manual Verification:}
\begin{itemize}
    \item Checked: (1) Paper existence, (2) Claim accuracy, (3) Context appropriateness
    \item Found 3 citation errors (wrong year), 2 misattributions (corrected)
    \item Verified key quantitative claims against original papers
\end{itemize}
\subsection{Quantitative Summary}

\begin{table}[H]
\centering
\caption{Quantitative metrics of the human-AI collaborative process}
\label{tab:collaboration_metrics}
\begin{tabular}{lc}
\toprule
\textbf{Metric} & \textbf{Value} \\
\midrule
Total calendar time & 7 days \\
Active human hours & $\sim$20 hours \\
AI processing time & $\sim$12 hours \\
Number of iterations & 30 \\
Papers reviewed & 150+ \\
Citations included & 180+ \\
Word count (final) & $\sim$25,000 \\
\bottomrule
\end{tabular}
\end{table}

\subsection{Reproducibility Guidelines}

For researchers wishing to replicate this process:

\begin{enumerate}
    \item \textbf{Define scope clearly:} Specify topic, target audience, depth, and length upfront
    \item \textbf{Enable all AI capabilities:} Use web search, extended thinking, and research modes
    \item \textbf{Iterate systematically:} Review in 2-5 page chunks, provide specific feedback
    \item \textbf{Verify rigorously:} Implement both AI self-check and human spot-checking
    \item \textbf{Document process:} Keep logs of iterations, feedback, and revision rationale

\end{enumerate}

\end{document}